\documentclass[a4paper,fleqn]{cas-dc}

\usepackage[utf8]{inputenc}
\usepackage{amsmath}
\usepackage{amsthm}
\usepackage{amssymb}
\usepackage{bm}
\usepackage{graphicx}
\usepackage{longtable}
\usepackage{booktabs}
\usepackage{circuitikz}
\usepackage{svg}
\usepackage{pdfpages}
\usepackage{float}
\usepackage[inline,shortlabels]{enumitem}
\usepackage{stfloats}
\usepackage{supertabular}
\usepackage{booktabs}
\usepackage{caption}
\usepackage[numbers, sort&compress]{natbib}
\usepackage{flushend}
\usepackage{tikz}
\usepackage{eurosym} 
\DeclareUnicodeCharacter{20AC}{\euro}
\usepackage{mathtools}
\usepackage{algorithm}
\usepackage{algpseudocode}
\usepackage{pgfplots}

\usepackage[nameinlink]{cleveref}
\crefname{figure}{Fig.}{Figs.}
\Crefname{figure}{Figure}{Figures}
\crefname{equation}{Eq.}{Eqs.}
\Crefname{equation}{Equation}{Equations}
\crefname{section}{Section}{Sections}
\crefname{table}{Table}{Tables}
\crefname{appendix}{Appendix}{Appendices}

\newcommand{\eg}{e.g.,\ }

\newcommand{\ie}{i.e.,\ }

\DeclareMathOperator*{\argmax}{\arg \max}

\newcommand{\state}{s}            
\newcommand{\enr}{e}              
\newcommand{\price}{c}            
\newcommand{\act}{u}              
\newcommand{\rew}{r}              
\newcommand{\dt}{\Delta t}        
\newcommand{\nsteps}{T}           
\newcommand{\Pmax}{P_{\max}}      
\newcommand{\Emax}{E_{\max}}      
\newcommand{\effc}{\eta_{c}}      
\newcommand{\effd}{\eta_{d}}      

\newcommand{\horizon}{H}          
\newcommand{\fan}{S}              
\newcommand{\scen}{\boldsymbol{\xi}} 
\newcommand{\scprob}{p}           

\newcommand{\tree}{\mathcal{T}}   
\newcommand{\Nodes}{\mathcal{N}}  
\newcommand{\Leaves}{\mathcal{L}} 
\newcommand{\rootn}{n_{0}}        
\newcommand{\nprob}{p}          
\newcommand{\cmean}{\bar{\price}} 
\newcommand{\nleaves}{L}          

\newcommand{\actp}{\act^{+}}      
\newcommand{\actn}{\act^{-}}      
\newcommand{\uvec}{\mathbf{\act}} 

\newcommand{\expcost}{C}          
\newcommand{\pathcost}{\Phi}      
\newcommand{\cvar}{\operatorname{CVaR}} 
\newcommand{\rlevel}{\beta}       
\newcommand{\rweight}{\lambda}    
\newcommand{\cvardual}{\alpha}    
\newcommand{\dualvec}{\boldsymbol{\cvardual}}
\newcommand{\lprob}{\tilde{\nprob}} 

\newcommand{\policy}{\pi_{\theta}}                
\newcommand{\polold}{\pi_{\theta_{\mathrm{old}}}} 
\newcommand{\gsize}{G}            
\newcommand{\token}{\mathbf{x}}   
\newcommand{\hid}{\mathbf{h}}     
\newcommand{\qvec}{\mathbf{q}}    
\newcommand{\logits}{\mathbf{z}}  
\newcommand{\cls}{\mathrm{CLS}}   
\newcommand{\nenc}{N}             
\newcommand{\asn}{a}              

\newcommand{\valf}{V_{\phi}}      
\newcommand{\ret}{G}              
\newcommand{\adv}{A}              
\newcommand{\ratio}{\rho}         
\newcommand{\clipeps}{\epsilon}   
\newcommand{\entcoef}{\omega}     

\newcommand{\MHA}{\operatorname{MHA}}
\newcommand{\FFN}{\operatorname{FFN}}
\newcommand{\MLP}{\operatorname{MLP}}
\newcommand{\LN}{\operatorname{LN}}
\newcommand{\GELU}{\operatorname{GELU}}
\newcommand{\softmax}{\operatorname{softmax}}
\newcommand{\clip}{\operatorname{clip}}

\begin{document}

\thinmuskip=1\thinmuskip
\medmuskip=1\medmuskip
\thickmuskip=1\thickmuskip

\title[mode = title]{Control-Oriented Scenario Tree Construction through Reinforcement Learning}
\shorttitle{Control-Oriented Scenario Tree Construction through RL}

\author[1]{Fabio Pavirani}[orcid=0009-0005-7904-099X]
\ead{fabio.pavirani@ugent.be}
\credit{Conceptualization, Methodology, Software, Formal analysis, Investigation, Visualization, Writing - Original Draft}

\author[2]{Bert Claessens}[orcid=0009-0006-6116-1483]
\credit{Conceptualization, Methodology, Supervision, Writing - Review \& Editing}

\author[3]{Pierre Pinson}[orcid=0000-0002-1480-0282]
\credit{Conceptualization, Supervision, Writing - Review \& Editing}

\author[1]{Chris Develder}[orcid=0000-0003-2707-4176]
\credit{Supervision, Project administration, Funding acquisition, Writing - Review \& Editing}

\affiliation[1]{organization={IDLab Ghent university -- imec},
                addressline={Technologiepark Zwijnaarde 126}, 
                postcode={9052}, 
                postcodesep={}, 
                city={Gent},
                country={Belgium}}
                
\affiliation[2]{organization={Beebop.ai},
                country={Belgium}}

\affiliation[3]{organization={Imperial College London},
                city={London},
                country={UK},}
                
\date{April 2024}

\shortauthors{Pavirani et al.}

\bibliographystyle{cas-model2-names}


\graphicspath{{pictures}}

\makeatletter\def\Hy@Warning#1{}\makeatother
\maketitle

\begin{abstract}
Multistage stochastic model predictive control (MPC) handles uncertainty by optimizing over a scenario tree, a finite branching approximation of future outcomes constructed from sampled forecasts. To build such a tree, conventional methods focus on matching the underlying probability distribution---e.g., via Wasserstein-based scenario reduction---but improved distributional accuracy does not necessarily yield better control performance.
We propose a control-oriented approach that learns scenario tree construction directly from its impact on downstream decisions. Fixing the tree topology, we formulate tree construction as a sequential assignment of sampled scenarios to leaves. This assignment is parameterized by an attention-based policy over the scenario set and trained using reinforcement learning, with closed-loop control profit as the objective. Training is stabilized by an asymmetric critic that leverages realized future trajectories.
We evaluate the method on a risk-averse battery arbitrage problem. Across a range of forecast set sizes, the learned construction consistently achieves the highest profit, outperforming classical forward and backward reduction methods and certainty-equivalent (single-trajectory forecast) control. The learned policy also exhibits greater robustness on challenging instances, consistently demonstrating better tail-risk characteristics.
Analysis of the resulting trees indicates that our method constructs compact, selectively branching structures that capture high-impact events while keeping most trajectories nearly deterministic. These findings highlight that the value of a scenario tree depends critically on the decisions it supports, and provide an effective framework to train scenario tree constructors merely based on the closed-loop control optimization signal.
\end{abstract}

\section{Introduction}
\label{sec:intro}
 
Model-based planning underpins decision-making across many engineering domains, and the operation of modern power systems is a prominent example. 
In such settings, decisions must be taken ahead of time based on predictions of uncertain future quantities, such as electricity prices or demand.

A common approach is to plan actions over a future horizon and commit to them in advance, allowing grid operators to schedule resources, balance supply and demand, and maintain network stability~\cite{morales2014integrating}. 
However, the value of such planning is increasingly limited by uncertainty. 
The growing share of renewable generation, volatile prices, and flexible demand introduce substantial variability~\cite{morales2014integrating}, so that decisions based on a single forecast can be far from optimal, or even infeasible, once reality deviates from it. 
This motivates planning techniques that explicitly account for uncertainty rather than reacting to it after the fact.
 
A principled way to address this challenge is multistage stochastic optimization, typically deployed in a receding-horizon or model predictive control (MPC) fashion~\cite{rawlings2017mpc,birge2011,mesbah2016stochastic}. 
Instead of optimizing against a single predicted trajectory, the controller considers multiple possible futures and optimizes decisions with respect to their expected performance (and, under risk aversion, their worst-case outcomes), while ensuring that decisions only depend on information available at the time they are taken. 
Repeated over time as new information arrives, this results in a closed-loop controller that anticipates uncertainty by construction and can be solved using standard optimization tools~\cite{shapiro2021}.
 
Applying this framework in practice requires approximating the continuous, high-dimensional distribution of future outcomes by a finite, computationally tractable object: a scenario tree. 
A scenario tree represents uncertainty as a set of branching trajectories, where each path corresponds to one possible future and branching points indicate when uncertainty is revealed and decisions can diverge. 
The tree therefore determines both the size of the optimization problem and how uncertainty is presented to the controller, making its construction central to control performance.
The dominant approaches build the tree by distribution matching; \ie scenarios are aggregated and reduced so as to minimize a probability metric\,---\,a Euclidean or, more generally, a Wasserstein-type distance\,---\,between the reduced tree and the original distribution~\cite{dupacova2003, heitsch2009, pflug2014}. These methods are well-founded and often effective, with stability guarantees that bound the deviation of the optimal value under the reduced tree.
 
Yet, \emph{a tree that approximates the distribution well is not necessarily a tree that yields good decisions}. 
Distribution-matching reduction is agnostic to the optimization problem the tree will feed. In other words, it minimizes a statistical distance but not the control cost, and there is no general guarantee that the scenarios it deems representative are the ones that matter most for the decision at hand~\cite{teichgraeber2019clustering}.
For example, two price trajectories that are close in their Euclidean space may call for opposite control actions, while two that are far apart may be handled identically; a reduction that is blind to the downstream objective cannot distinguish these cases. 
This gap between distributional fidelity and decision quality has motivated a broader line of work on \emph{decision-focused} or task-based learning, in which models and approximations are trained against the downstream objective rather than a surrogate loss~\citep{donti2017, elmachtoub2022}, and on problem-driven scenario generation that tailors the reduction to the optimization at hand~\citep{fairbrother2022}.
 
In this work, we take a data-driven, control-oriented view of scenario-tree construction. 
Instead of optimizing a statistical distance, we directly optimize the quality of the decisions induced by the tree using reinforcement learning (RL). 
Concretely, we fix the tree topology and cast its construction as a sequential assignment problem: given a set of sampled future trajectories, the model learns how to assign them to the leaves of the tree so as to maximize the resulting closed-loop control performance. 
The policy is parameterized by an attention-based neural network and trained with Proximal Policy Optimization (PPO)~\citep{schulman2017proximal,lee2019set}. 
The resulting objective is therefore the realized control profit, rather than any proxy measure of distributional accuracy.
 
We study the approach on a battery energy-arbitrage problem, a canonical storage-control task in which the operator charges and discharges against uncertain prices under a risk-averse objective. 
Benchmarking against \begin{enumerate*}[(i)]
    \item classical reduction heuristics (forward selection, backward reduction), 
    \item a random-assignment baseline on the same topology, 
    \item a certainty-equivalent (deterministic) controller, and
    \item an upper-bound unrealistic perfect-foresight oracle, 
\end{enumerate*}
our learned policy attains the highest profit among all realistic methods (\ie all but the oracle one).
The learned policy's advantage is largest in the small-sample regime that keeps the online optimization tractable, and it narrows as the forecast is sampled more densely and certainty-equivalent control approaches optimality for this linear problem\,---\,a crossover that the learned policy reaches last, and from above, never falling below the certainty-equivalent reference (as the distance-based heuristics do). 
An analysis of the learned trees shows that the policy spends most of its decisions building compact, near-deterministic trees and branches only selectively, which both explains its behavior and keeps the solved programs (\ie the complexity of the optimization problem) limited.
 
Our contributions are:
\begin{itemize}
  \item We formulate a scenario tree construction framework for multistage stochastic MPC, defining it as a sequential decision problem. The framework allows a control-oriented model that learns the scenario tree construction with reinforcement learning, directly optimizing the downstream realized control cost rather than a distributional distance.
  \item We design and evaluate an attention-based assignment policy that maps an unordered set of forecast scenarios onto the leaves of a fixed tree topology, together with an asymmetric critic that exploits privileged future information at training time only.
  \item On a risk-averse battery-arbitrage task, we show that the learned construction outperforms classical distance-based reduction and certainty-equivalent control, characterize the regime in which learning pays off, and analyze the structure of the compact and selectively branched trees the policy produces. 
\end{itemize}
 
\section{Related Work}
\label{sec:related}
 
The construction of scenario trees for multistage stochastic programming is classically posed as an approximation of the underlying probability distribution.
Reduction methods select a small set of representative scenarios that minimize a probability metric to the original distribution, with forward-selection and backward-reduction heuristics being the standard algorithms~\citep{dupacova2003, heitsch2009}; the theory extends to multistage problems through process distances such as the nested distance~\citep{pflug2014}, and recent work has sharpened the fundamental limits and guarantees of distance-based reduction~\citep{rujeerapaiboon2022}. These methods come with stability results that bound how much the optimal value can change under the reduced tree, and they are the workhorses against which we compare. Their objective, however, is distributional fidelity: they minimize a statistical distance, with no reference to the decision the tree will inform. Our method departs from this premise, optimizing the tree directly against the realized control cost rather than a distance to the forecast.
 
The latter concern is not new. A line of work close to ours recognizes that distributional accuracy does not necessarily align with decision quality, and tailors the scenario reduction steps to the optimization problem itself, measuring scenario similarity in a \emph{problem space} induced by the decisions the scenarios entail rather than in the sample space directly~\citep{fairbrother2022}. Within this line, \citet{bertsimas2023optimization} cast two-stage scenario reduction as a convex optimization that selects a representative distribution preserving the quality of the resulting sample-average decision, with recovery guarantees as the sample size grows. \citet{hewitt2022decision} formalize the same intuition through a symmetric opportunity-cost notion of similarity, grouping scenarios by mutual decision impact via a graph-clustering heuristic, while \citet{keutchayan2023problem} pursue a related problem-driven clustering scheme with an asymmetric distance. Closest to our domain, \citet{zhuang2025problem} make the idea concrete for two-stage stochastic dispatch in power systems: scenarios are projected onto a problem space defined by the dispatch program, an opportunity-cost-based distance is defined between them, and the representative set is selected by a single mixed-integer program; in a following work~\cite{zhuang2026iterative} the same authors extended the framework to risk-averse formulations with a CVaR objective through an iterative variant, in the same spirit as the earlier scheme of \citet{garcia2014iterative}, who iteratively refine the tail-scenario set against a CVaR criterion.
These approaches share our motivation, but differ in mechanism and scope. They are one-shot or iterative procedures, typically derived for two-stage formulations, and they construct the representative set by solving an auxiliary optimization problem against the current forecast\,---\,a \emph{per-instance} reduction whose offline cost grows with the original sample size and that must be repeated whenever the forecast or the system state changes. 
We instead learn a purely data-driven amortized construction policy\,---\,trained once and reused at every control step\,---\,whose training signal is the realized closed-loop objective cost/reward of a multistage, receding-horizon controller, obtained by simply rolling out the controller rather than by formulating a tractable surrogate of the reduction problem to be solved at decision time.
 
The same decision-oriented principle we use has been pursued from the machine-learning side, under the banner of decision-focused or task-based learning, where a predictive model is trained so that its outputs lead to good downstream decisions rather than low prediction error. This is achieved by differentiating through the optimization layer~\citep{donti2017, amos2017optnet, wilder2019} or through surrogate losses such as the smart predict-then-optimize loss~\citep{elmachtoub2022}. 
Two distinctions set our setting apart. First, we do not learn the forecast: the predictive distribution is given as a fan of samples, and we learn how to \emph{reduce} that fan into a tree.
Second, building a tree from samples is a discrete, combinatorial operation\,---\,assigning scenarios to branches and pruning the result\,---\,through which gradients do not flow; rather than relax or differentiate through it, we treat the construction as a sequential decision process and optimize it with reinforcement learning, whose reward is the multistage control cost itself.

Casting the problem this way places it alongside a growing body of work that embeds learning inside optimization pipelines\,---\,\eg learning to branch or select cuts in mixed-integer solvers~\citep{gasse2019exact}, and solving combinatorial problems with attention-based policies trained by policy gradient~\citep{vinyals2015pointer,
kool2019attention}. Our assignment policy architecturally fits in this lineage: it reads an unordered set with attention and emits an autoregressive sequence of discrete assignments trained with a policy-gradient method. 
What we learn is nonetheless different, in that we shape the uncertainty representation consumed by an exact optimizer rather than learning to solve or accelerate the optimization. 
The same holds for hybrid schemes that combine reinforcement learning with MPC by learning cost or value functions to steer the controller~\citep{gros2020data,madahi2025model}: we leave the controller's objective untouched and learn only the scenario tree it acts on.
 
This last distinction is sharpest against model-free RL, which has been applied directly to battery and storage dispatch, replacing the optimizer with a learned policy that maps a state to an action~\citep{cao2020deep,madahi2024distributional}. 
Such controllers are flexible but must learn to respect operational constraints\,---\,\eg state-of-charge limits, power bounds, and risk preferences\,---\,from reward signals alone, which is notoriously difficult and offers no feasibility guarantees. 
Our design is deliberately the opposite: the exact, constrained, risk-averse multistage program remains in the loop and continues to guarantee feasibility and to encode the risk objective, while reinforcement learning is confined to constructing the uncertainty input the program reasons over. We thus aim to combine the adaptivity of reinforcement learning with the rigor of mathematical optimization, rather than substituting one for the other.
 
Taken together, these threads locate our contribution at their intersection. 
To our knowledge, this is the first method to learn multistage scenario-tree construction as a set-to-assignment policy trained, by reinforcement learning, on the realized closed-loop cost of the downstream controller, while retaining an exact risk-averse optimizer. It inherits the decision-oriented goal of problem-driven scenario reduction but replaces one-shot, distance-based procedures with a learned and reusable policy; it shares the end-to-end objective of decision-focused learning but targets a non-differentiable reduction step through reinforcement learning rather than the forecast through differentiable optimization; finally it borrows the architecture of neural combinatorial optimization while applying it to the representation of uncertainty rather than to the solution of the optimization.

\section{Problem Formulation}
\label{sec:problem}
 
We consider a grid-connected battery that performs energy arbitrage, charging when
electricity is cheap and discharging when it is expensive. We first formalize the
dispatch problem that an operator would solve if the price trajectory over the
planning horizon were known (\cref{sec:dispatch}), and then extend it to the
realistic setting in which only a probabilistic forecast is available. The latter
is handled by a multistage stochastic program defined on a scenario tree
(\cref{sec:tree}), the construction of which is the central object of this
work.
 
\subsection{Battery arbitrage dispatch}
\label{sec:dispatch}

We consider a simple but representative control problem: operating a battery to perform energy arbitrage. 
The objective is to charge the battery when electricity prices are low and discharge it when prices are high, while respecting physical constraints such as capacity limits and maximum charging power.

We consider a planning horizon of $\horizon$ steps of length $\dt$ hours. At each timestep $\tau=0, \dots, \horizon-1$, the battery is described by its stored energy $\enr_\tau \in [0,\Emax]$ and is operated through a net
charging power $\act_\tau = \actp_\tau - \actn_\tau$, decomposed into nonnegative charging
and discharging components $\actp_\tau, \actn_\tau \ge 0$, with $\act_\tau > 0$ denoting
charging and $\act_\tau < 0$ discharging. Over one step the stored energy evolves as
\begin{equation}
  \enr_{\tau+1} \;=\; \enr_\tau + \left(\effc\,\actp_\tau - \frac{\actn_\tau}{\effd}\right)\,\dt,
  \qquad \tau = 0,\dots,\horizon-1,
  \label{eq:dynamics}
\end{equation}
where $\effc,\effd \in ]0,1]$ are the charge and discharge efficiencies, and the
state and control are subject to the operating limits
\begin{equation}
  0 \le \enr_\tau \le \Emax,
  \qquad
  0 \le \actp_\tau,\,\actn_\tau \le \Pmax .
  \label{eq:limits}
\end{equation}
 
The energy exchanged with the grid over step $\tau$ is defined as $\act_\tau\,\dt$. Given a price trajectory $\price_{0:\horizon-1}$, the instantaneous arbitrage profit is then
\begin{equation}
  \rew_\tau \;=\; -\,\price_\tau\,\act_\tau\,\dt .
  \label{eq:reward}
\end{equation}
Starting from a given initial energy $\enr_0$, the optimal dispatch over the
horizon solves the linear program
\begin{equation}
\label{eq:dispatch}
    \begin{aligned}
        \underset{\{\act_\tau\}_{\tau=0, \dots, \horizon-1}}{\max}\quad
        & \sum_{\tau=0}^{\horizon-1}
          -\,\price_\tau\,\act_\tau\,\dt
        \\
        \text{subject to}\quad
        & \eqref{eq:dynamics}\text{--}\eqref{eq:limits}.
    \end{aligned}
\end{equation}
 
In closed loop, the operator applies the first optimal action in a receding-horizon fashion\,---\,solving~\eqref{eq:dispatch} over the window $[t, t+\horizon[$, applying $\act_t$, advancing one step, and re-solving\,---\,so as to maximize the realized cumulative profit $\sum_{t=0}^{\nsteps} \rew_t$. 
Were the future prices known, this loop would be optimal; \ie given a long-enough horizon $\horizon$, solving~\eqref{eq:dispatch} with the realized prices yields the perfect-foresight optimum, an unattainable upper bound we use as a reference in \cref{sec:setup}. 
In practice the future price trajectory is uncertain, which motivates the stochastic formulation below.
 
\subsection{Decision-making under uncertainty with scenario trees}
\label{sec:tree}
 
At decision time, the operator does not observe future prices but only a probabilistic forecast, represented by a fan of $\fan$ sampled price trajectories $\left[\scen^{(i)}\right]_{i=1}^{\fan}$ over the horizon $\horizon$, with $\scen^{(i)}\in\mathbb{R}^H$ each carrying a probability $\scprob_i$. 
Replacing the unknown prices in~\eqref{eq:dispatch} by a single point forecast (\eg a trajectory containing the average of all the sampled prices) yields a certainty-equivalent controller that disregards uncertainty altogether. 
Properly accounting for uncertainty calls instead for a multi-stage stochastic program, in which the dispatch may adapt as prices are progressively revealed over the horizon, while satisfying the non-anticipativity property (\ie the decision taken at any stage may depend only on the information available up to that stage). 
Formulating such a program directly over all $\fan$ trajectories usually is computationally prohibitive when $\fan$ is large. Therefore, the uncertainty is approximated by a scenario tree that aggregates the trajectories in a compact structure.
 
A scenario tree $\tree$ organizes the probabilistic forecast into a branching structure that reflects how uncertainty unfolds across the horizon. 
Its root $\rootn$ represents the present step at which the state is known and a single decision must be taken; each subsequent child node corresponds to the decisions/states following the current one, and the branches leaving a node enumerate the alternative ways the future may evolve from the information state that node encodes. 
A node $n$ thus aggregates the scenarios that remain indistinguishable up to its stage, the root-to-leaf paths through the leaf set
$\Leaves$ are the representative trajectories the tree retains, and the branching points mark the stages at which the controller must commit to a decision before the associated uncertainty is resolved. 
Relative to the full fan of sampled trajectories (\ie the probabilistic forecast), a tree with few nodes yields a small and tractable optimization problem, at the price of approximating the forecast distribution\,---\,an approximation whose fidelity depends on which scenarios are retained and where the tree is allowed to branch.
 
The dispatch model of \cref{sec:dispatch} is lifted onto the tree without modification. 
Every node $n$ carries its own energy state and charging decision; the energy recursion (\cref{eq:dynamics}) is imposed along each parent-to-child edge, relating a node's state to that of its children, and the operating limits (\cref{eq:limits}) together with a terminal feasibility condition hold at every node. 
Non-anticipativity is naturally enforced: scenarios that share a node must also share the same decision up to that point, ensuring that decisions only depend on currently available information.
Each node is summarized by the probability mass and conditional mean price of the scenarios it aggregates,
\begin{equation}
  \nprob_n \;=\; \sum_{i \in n} \scprob_i,
  \qquad
  \cmean_n \;=\; \frac{1}{\nprob_n}\sum_{i \in n} \scprob_i\,\scen^{(i)}_{t_n},
  \label{eq:nodestats}
\end{equation}
where $t_n$ is the look-ahead index of node $n$.
 
The deterministic objective~\eqref{eq:dispatch} generalizes to a
probability-weighted expectation over the tree, augmented with a risk term that
penalizes costly branches. The expected dispatch cost is
\begin{equation}
  \expcost(\uvec) \;=\; \sum_{n \in \Nodes} \nprob_n\,\cmean_n\,\act_n\,\dt,
  \label{eq:expcost}
\end{equation}
where $\Nodes$ denotes all the nodes in the tree. 
In addition to optimizing expected performance, we account for risk by penalizing unfavorable outcomes. 
To this end, we follow~\cite{zhuang2026iterative,garcia2014iterative} and use Conditional Value-at-Risk (CVaR), which captures the expected cost in the worst-performing scenarios and allows the controller to be sensitive to rare but impactful events.
By associating with each leaf $\ell \in \Leaves$ the cost accumulated along its root-to-leaf path: 
\begin{equation}
    \pathcost_\ell(\uvec) = \sum_{n \preceq \ell} \cmean_n\,\act_n\,\dt \;,
\end{equation}
we can control the tail risk through a weighted sum of CVaR terms at the confidence level $\rlevel$. 
Using the Rockafellar--Uryasev
representation~\cite{rockafellar2000},
\begin{equation}
  \cvar_{\rlevel}(\uvec, \cvardual)
  \;=\;
  \cvardual
  + \frac{1}{1-\rlevel}\sum_{\ell \in \Leaves}
      \lprob_\ell\,\big[\pathcost_\ell(\uvec) - \cvardual\big]^{+},
  \label{eq:cvar}
\end{equation}
with $\lprob_\ell$ the leaf probabilities, the controller solves\footnote{$\uvec$ indicates the vector of actions for each node of the tree. $\dualvec$ indicates the vector of CVaR values $\left[ \cvardual_k \right]_k$}
\begin{equation}
\label{eq:mpc}
    \begin{aligned}
        \max_{\uvec,\,\dualvec}\quad
        & -\Big(
            \expcost(\uvec)
            + \sum_k \rweight_k\,\cvar_{\rlevel_k}(\uvec, \cvardual_k)
          \Big)
        \\
        \text{s.t.}\quad
        & \text{the per-node form of \cref{eq:dynamics,eq:limits},}
    \end{aligned}
\end{equation}
with multiple CVaR terms $\cvardual_k, \rlevel_k, \rweight_k$ (iterated with a generic index $k$). We use several confidence levels to allow the controller to shape different parts of the loss tail rather than penalizing only a single quantile region. 
\Cref{eq:mpc} remains a linear program once the CVaR terms are linearized
through \cref{eq:cvar}. As in the deterministic case, only the root decision
$\act_{\rootn}$ is implemented before the horizon recedes and the program is
re-solved at the next step.
 
The formulation defined in \cref {eq:nodestats,eq:expcost,eq:cvar,eq:mpc} is common to every method
studied in this paper. What it leaves unspecified is the construction of the tree
$\tree$ from the fan\,---\,which scenarios to retain, how to group them, and where to
branch. This choice fixes both the size of the program~\eqref{eq:mpc} and how
faithfully it represents the forecast, and is the subject of the remainder of the
paper.

\begin{figure*}[h]
    \centering
    \includegraphics[width=\linewidth]{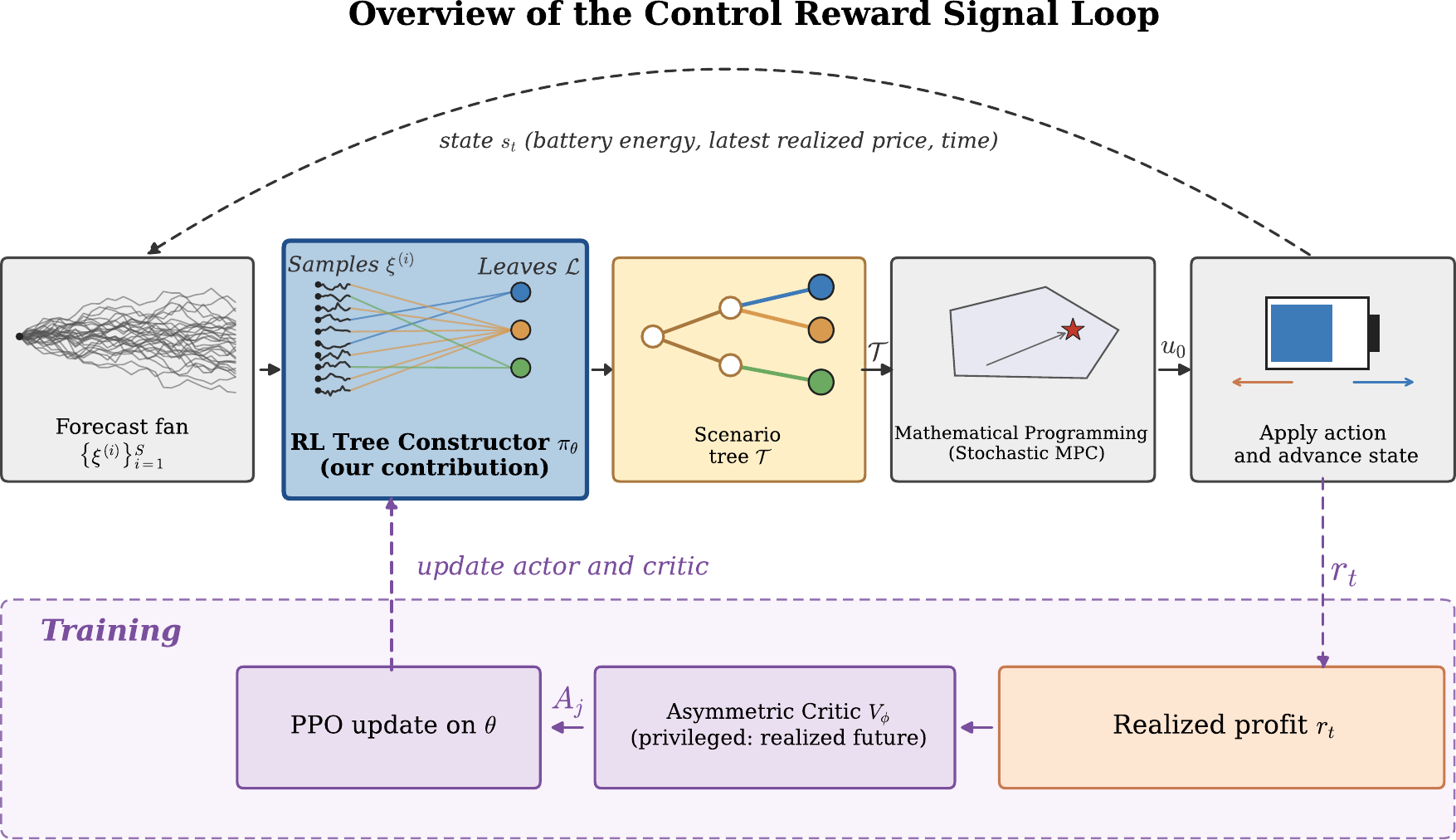}
    \caption{Control Loop used in our training framework. First, a fan of price trajectories is generated from a random distribution. Then, each trajectory gets assigned to a tree leaf with a fixed topology by a differentiable actor, constructing a scenario tree. The scenario tree is then fed into a mathematical programming solver, which applies the upcoming action to a battery, updating the system state and repeating the loop in a receding-horizon fashion. The constructor agent (made out of an actor and a critic, as in standard PPO) gets trained directly on the closed-loop control profit.}
    \label{fig:control_overview}
\end{figure*}

\section{Methodology}
\label{sec:method}
 
The framework of \cref{sec:problem} leaves open how the scenario tree
$\tree$ is built from the fan. In our work, we learn this construction with a data-driven model. 
We frame the construction as a short sequential assignment problem (\cref{sec:assignment}), solve it with an attention-based policy and an asymmetric critic (\cref{sec:arch}), and train them jointly with reinforcement learning so that the sole objective is the optimization of the control signal (\ie the maximization of the reward, of the minimization of the cost) (\cref{sec:training}). An overview of the control loop used to train and evaluate the agent is depicted in \cref{fig:control_overview}

\subsection{Building the tree by sequential assignment}
\label{sec:assignment}

Our policy does not design the shape of the tree; it decides how the sampled scenarios populate a tree whose shape is fixed in advance. 
We therefore start from a fixed \emph{topology}: an empty template that specifies a root, a set of $\nleaves$ leaves, and the branches (node trajectories) that connect them, but that holds no scenarios yet. 
Since each leaf is reached from the root by a unique path, this template already fixes which intermediate nodes are shared between which leaves; the only thing left undecided is which scenarios travel down which branch.

\begin{figure}
    \centering
    \includegraphics[width=0.5\textwidth]{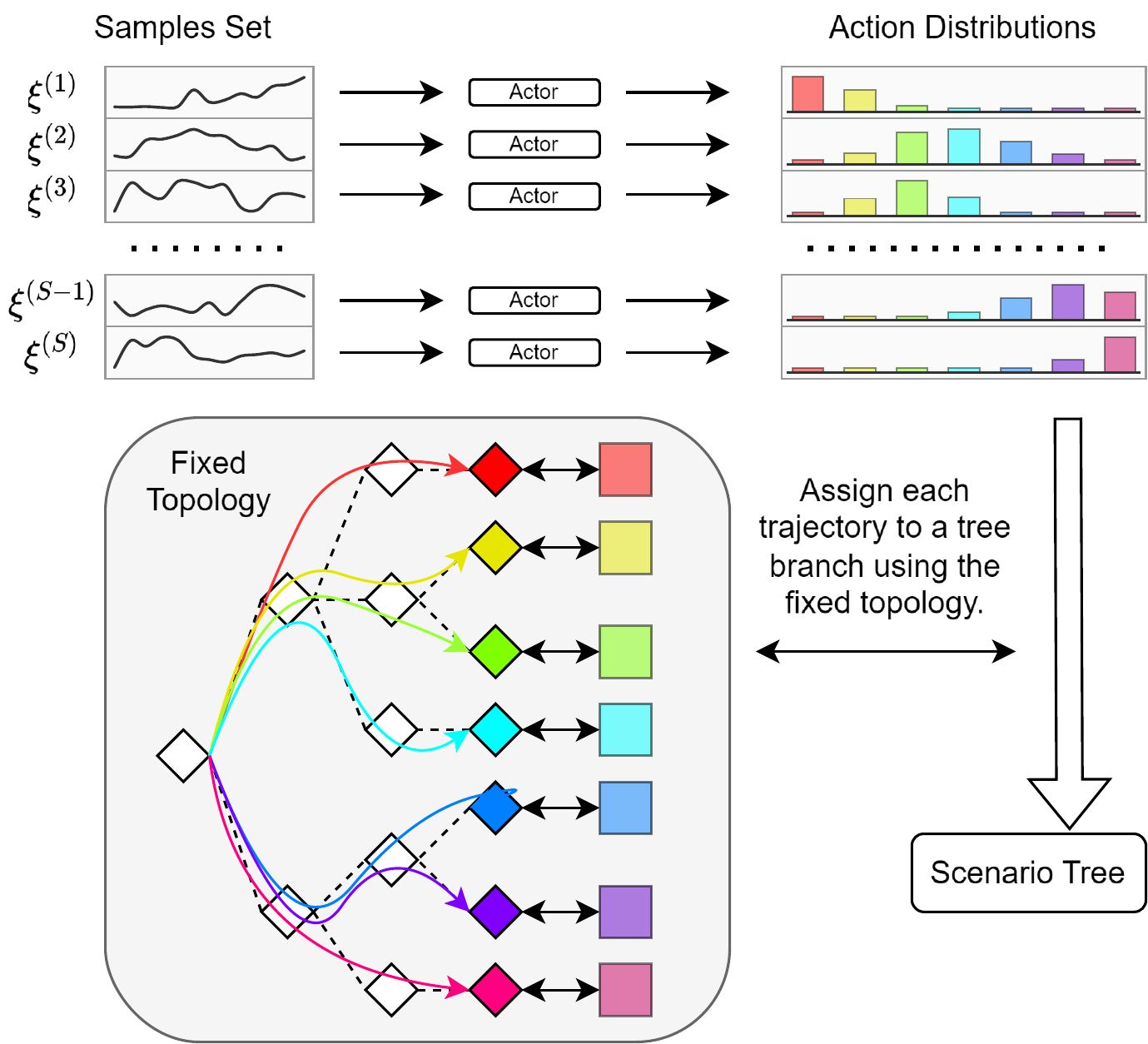}
    \caption{Graphical illustration of the scenario tree construction through
    assignment. Each sampled trajectory $\scen^{(i)}$ is processed by a
    differentiable actor to obtain an assignment probability distribution,
    giving the probability of assigning each sample to a given branch of a fixed
    tree topology, indicated using different colors. During training, the tree
    is built by sampling from these assignment distributions; during inference,
    the assignments are obtained through $\argmax$.}
    \label{fig:scenario_leaf_assignment}
\end{figure}

Filling the template comes down to a single operation of assigning each scenario to one leaf, repeated over all scenarios. 
Assigning scenario $i$ to a leaf places it in every node along the path from the root to that leaf.
Two scenarios sent to the same leaf therefore share all the nodes their paths have in common, and scenarios sent to different leaves are treated as identical up to the stage at which those paths separate\,---\,this is the non-anticipativity property discussed in \cref{sec:problem}, now realized concretely by the assignment.
Once every scenario has been placed, the tree is fully populated: each node knows the set of scenarios routed through it, and its probability mass and representative price are obtained by aggregating that set as in \cref{eq:nodestats}.
Any leaf that receives no scenario, together with any branch that is thereby left empty, is pruned.
The policy thus implicitly decides not only how scenarios cluster but also how many branches the final tree really uses; in the extreme, routing all scenarios to a single leaf collapses the tree to one trajectory and reduces the controller to the certainty-equivalent dispatch of \cref{sec:problem}.
A graphical illustration of the scenario-leaf assignment is depicted in \cref{fig:scenario_leaf_assignment}.

Rather than placing all $\fan$ scenarios in one shot, the policy assigns them a few at a time (\ie groups the fan and assigns one group at a time), so that each decision can take the placements already made into account.
This requires each partial construction to carry a state, which is represented
through individual features built during the construction.
Specifically, each scenario is described by a feature vector\,---\,its \emph{token}, defined
formally in \cref{sec:arch}\,---\,which, beyond encoding the scenario's price
trajectory and system state (\ie the state of the control system to be
optimized; in our case the state of the battery), holds two bookkeeping fields:
the leaf it has been assigned to so far (encoded with a one-hot representation, with an extra ``unassigned'' entry), 
and a flag indicating whether it is one of the scenarios being decided at the current group step.

Given a timestep $t$, the corresponding scenarios are arranged in a fixed order, by decreasing Euclidean deviation from the mean trajectory, so that the most atypical scenarios are placed first.
They are then split into $K = \left\lceil \frac{\fan}{\gsize} \right\rceil$ groups (each spanning $\gsize$ scenarios or fewer), denoted by $\mathcal{G}_t^{(k)} \subseteq \{\scen^{(i)}\}_{i=1}^{\fan}$ with:
\begin{equation}
    |\mathcal{G}_t^{(k)}| \le \gsize \;,\; \forall k=1,\dots,K \;.
    \label{eq:groups}
\end{equation}
Construction proceeds over group steps $k=1,\dots,K$ for each timestep $t$.
At group step $k$, the policy reads the tokens of all $\fan$ scenarios\,---\,so
it sees the whole fan and the partially built tree simultaneously\,---\,and
assigns only the scenarios in the current group $\mathcal{G}_t^{(k)}$ to the corresponding fixed topology leaves.
We denote the corresponding group action by
\begin{equation}
    \asn_t^{(k)}
    =
    \left(
        \asn_t^{(i,k)}
    \right)_{i\in\mathcal{G}_t^{(k)}} \;;
    \qquad
    \asn_t^{(i,k)} \in \{1,\dots,\nleaves\}.
    \label{eq:groupaction}
\end{equation}
After this action is taken, the tokens of the scenarios in $\mathcal{G}_t^{(k)}$ are
updated to record the chosen leaves, the next group becomes current, and the
step repeats until no scenario remains unassigned.

\begin{figure*}
    \centering
    \includegraphics[width=\linewidth]{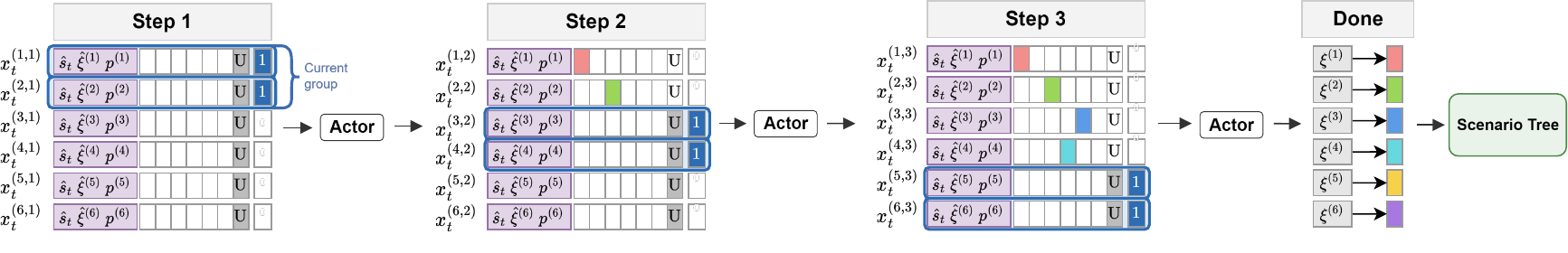}
    \caption{Example showing the sequential steps the actor takes during one
    scenario tree construction, in a case with $\fan=6$ and $\gsize=2$.
    The scenarios are split into 3 groups that get assigned
    sequentially. After each group action, the corresponding assignment tokens
    are updated, with colors representing branches as in
    \cref{fig:scenario_leaf_assignment}, until all scenarios are assigned.}
    \label{fig:sequential_assignment}
\end{figure*}

The state of the construction from one group step to the next is passed through two bookkeeping state fields: the \emph{assigned-leaf} field reports the tree built so far, and the \emph{current-group} field tells the policy which scenarios it must place now.
This iteration is depicted in \cref{fig:sequential_assignment}.

Constructing one tree is therefore a short sequential decision process, which we
solve with reinforcement learning.
Each action assigns one group of scenarios to leaves; after the final group, the
completed tree is handed to the optimization~\eqref{eq:mpc}, the first-stage
action is applied, and the environment returns the arbitrage
profit~\eqref{eq:reward}.
We use this profit as the reward, so the policy is optimized directly for
downstream control performance, rather than for any statistical resemblance
between the tree and the fan (as shown in \cref{fig:control_overview}).

\subsection{Network architecture}
\label{sec:arch}

The assignment policy acts on a set of $\fan$ scenarios, and the leaf chosen for
any scenario should depend on the rest of the fan and on the assignments already
committed, not on that scenario in isolation.
We therefore parameterize the policy with a cross-attention set encoder, which
is permutation-equivariant over the fan and conditions each scenario's
representation on all the others.
The value function is parameterized by a separate network of the same design;
\ie the actor and critic are trained jointly but share no parameters.

At group step $k$ in timestep $t$, each scenario is represented by a token vector
\begin{equation}
  \token_t^{(i,k)}
  =
  \big[
    \hat{\state}_t,\;
    \hat{\scen}^{(i)},\;
    \scprob^{(i)},\;
    b^{(i, k)},\;
    m^{(i, k)}
  \big],
  \label{eq:token}
\end{equation}
concatenating the normalized system state $\hat{\state}_t$, the scenario's normalized price trajectory $\hat{\scen}^{(i)}$ over the horizon, its probability
$\scprob_i$, and two features describing the partial tree construction.
The vector $b^{(i, k)}$ is the one-hot encoding of the leaf currently assigned to
scenario $i$ before processing group $\mathcal{G}_t^{(k)}$, with an additional entry
for unassigned scenarios.
The indicator
\begin{equation}
    m^{(i, k)}
    =
    \mathbb{1}_{\mathcal{G}_t^{(k)}}(i)
    \label{eq:currentgroupindicator}
\end{equation}
marks whether scenario $i$ belongs to the group currently being assigned.
As the network observes a scenario only through its token, these two fields are
the channel through which the partially constructed tree is exposed to the
policy.

At each group-assigment step $k$, each token is first independently embedded into a latent vector using a shared projection (\ie trained parameters $W$):
\begin{equation}
  \hid_t^{(i, k)}
  =
  \LN\!\big(\GELU(W\,\token_t^{(i, k)})\big),
  \qquad
  i=1,\dots,\fan .
  \label{eq:tokenembedding}
\end{equation}
with $\LN(\cdot)$ denoting layer normalization.
This produces the context set
\begin{equation}
    C^{(k)}
    =
    \left\{\hid_t^{(i, k)}\right\}_{i=1}^{\fan},
    \label{eq:contextset}
\end{equation}
which represents all scenarios together with the partial assignment state before
processing group $\mathcal{G}_t^{(k)}$.
Within a given group step, these context vectors are kept fixed throughout the
cross-attention encoder and provide information about the whole fan of
scenarios.

Next, we process the scenarios using multiple multi-head attention layers. 
Instead of processing all scenarios jointly as queries, the network
focuses only on the scenarios that are currently being assigned.
To do this, it builds a query set containing:
\begin{enumerate*}[(i)]
    \item a learnable global token $\qvec_{\cls}$, and
    \item the embeddings of the scenarios in the current group
    $\left\{\hid_t^{(i, k)} : i \in \mathcal{G}_t^{(k)}\right\}$.
\end{enumerate*}
The initial query set is therefore
\begin{equation}
    Q_t^{(k,0)}
    =
    \left[
        \qvec_{\cls},\;
        \{\hid_t^{(i, k)} : i \in \mathcal{G}_t^{(k)}\}
    \right].
    \label{eq:queryset}
\end{equation}
The query set is then updated using $\nenc$ layers of cross-attention.
For layer $s=0,\dots,\nenc-1$, the queries attend to the full context
$C^{(k)}$, \ie to all scenarios:
\begin{equation}
    \begin{aligned}
      \widetilde{Q}_t^{(k,s+1)}
      &=
      Q_t^{(k,s)}
      +
      \MHA\!\left(
        \LN(Q_t^{(k,s)}),
        \LN(C^{(k)}),
        \LN(C^{(k)})
      \right), \\
      Q_t^{(k,s+1)}
      &=
      \widetilde{Q}_t^{(k,s+1)}
      +
      \FFN\!\left(\LN(\widetilde{Q}_t^{(k,s+1)})\right).
    \end{aligned}
    \label{eq:crossattention}
\end{equation}
Here $\MHA(\cdot, \cdot, \cdot)$ denotes multi-head attention and $\FFN(\cdot)$ a
feed-forward network.

Intuitively, each query vector repeatedly looks at all scenarios and extracts
the information it needs.
After $\nenc$ layers, the refined query corresponding to each current scenario
encodes how that scenario should be interpreted relative to the entire fan and
the current partial assignments.
This architecture avoids full pairwise interactions between all scenarios.
Instead, only the current group attends to the full set, leading to a
computational cost of $O(\gsize \cdot \fan)$ rather than $O(\fan^2)$.

For each current scenario $i \in \mathcal{G}_t^{(k)}$, let
$\qvec_t^{(i, k)}$ denote the final query representation associated with that
scenario.
The actor maps this representation through a multilayer perceptron (MLP) to
produce logits over the leaves $\ell = 1, \dots, \nleaves$:
\begin{equation}
\begin{aligned}
  &\logits_t^{(i,k)}
    = \MLP\!\big(\qvec_t^{(i,k)}\big), \\
  &\policy(\asn_t^{(i,k)} = \ell \mid \tree_t^{(k)})
    = \softmax\!\big(\logits_t^{(i,k)}\big)_\ell .
\end{aligned}
\label{eq:actorreadout}
\end{equation}
Here $\tree_t^{(k)}$ denotes the tokenized partial construction state at group step
$k$, namely
\begin{equation}
    \tree_t^{(k)}
    :=
    \left\{ \token_t^{(i, k)} \right\}_{i=1}^{\fan}.
    \label{eq:tokenizedstate}
\end{equation}
Conditioned on this shared state representation, the group assignment
distribution factorizes over the scenarios in the current group:
\begin{equation}
    \policy\left(\asn_t^{(k)} \mid \tree_t^{(k)}\right)
    =
    \prod_{i\in\mathcal{G}_t^{(k)}}
    \policy\left(\asn_t^{(i, k)} \mid \tree_t^{(k)}\right).
    \label{eq:groupfactorization}
\end{equation}
During training, assignments are sampled from this distribution, while at evaluation time the most probable leaves are selected.

\begin{figure*}
    \centering
    \includegraphics[width=\linewidth]{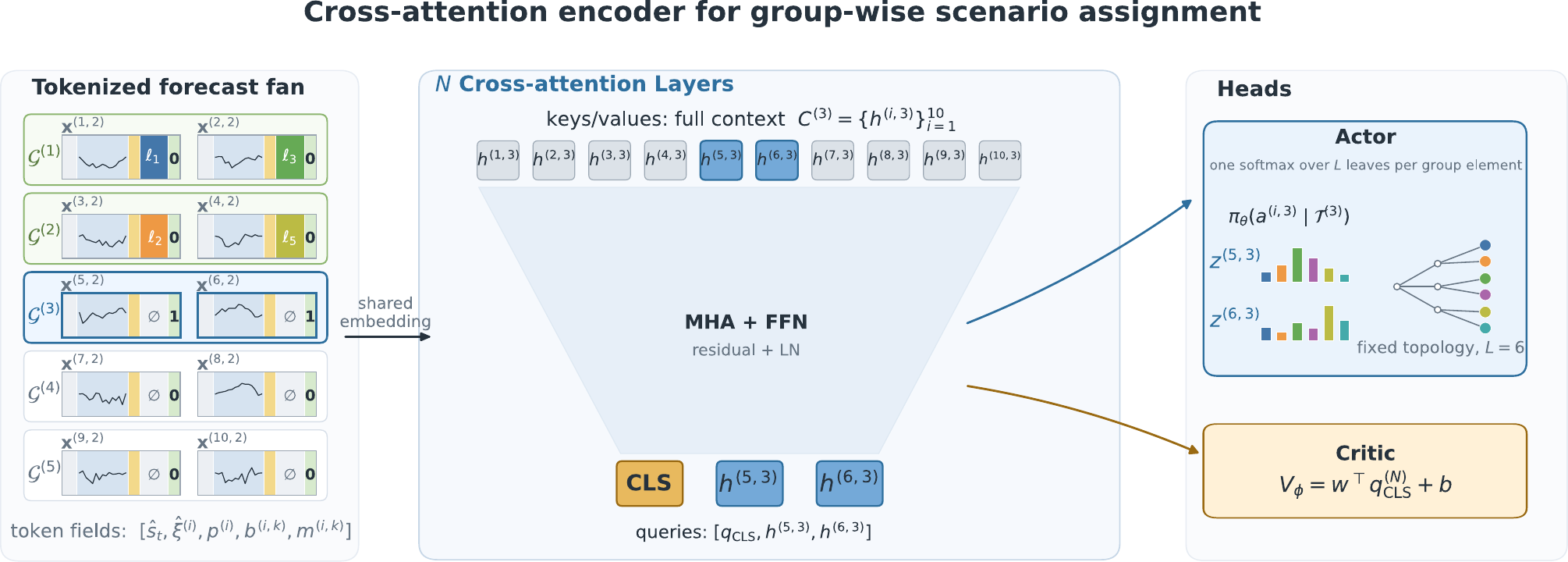}
    \caption{Architecture visualization of a construction step, with $\gsize=2$, and $\fan=10$. The example cover the forward step with $k=3$, meaning that groups 1-2 have been assigned already, group 3 is currently under assignment, and groups 4-5 are left to be assigned in the following steps.}
    \label{fig:architecture_figure}
\end{figure*}

The critic uses the same cross-attention encoder as the actor, but with an independent set of parameters.
In particular, it similarly constructs a context $C^{(k)}$ and query set $Q^{(k,0)}$, consisting of the learnable token $\qvec_{\cls}$ and the current scenarios (\ie the scenarios in the current assignment group), and applies the same sequence of cross-attention layers described above.
The key difference lies in the readout.
Instead of producing per-scenario action logits, the critic maps the final
representation of the global query token $\qvec_{\cls}^{(k,\nenc)}$ to a scalar
value estimate:
\begin{equation}
  \valf(\tree_t^{(k)})
  =
  w^\top \qvec_{\cls}^{(k,\nenc)} + b .
  \label{eq:criticreadout}
\end{equation}
This yields a prediction of the value of the current partial construction state.
In addition, the critic has access to privileged information during training:
its input tokens are augmented with the realized future trajectories over the
horizon.
This information is not available to the actor and is used solely to reduce the
variance of the value estimate.
At deployment time, only the actor is used, so the dependency on future realized trajectories gets dropped, as in realistic conditions.
The full architecture is depicted in \cref{fig:architecture_figure}

\subsection{Training with PPO}
\label{sec:training}

We train with Proximal Policy Optimization (PPO)~\cite{schulman2017proximal}.
An episode is one receding-horizon rollout of length $\nsteps$.
At each physical time step $t$, the policy constructs one scenario tree through
$K=\lceil \fan/\gsize \rceil$ group decisions, indexed by $k=1,\dots,K$.
Each pair $(t,k)$ is stored as a transition, together with the partial
construction state $\tree_t^{(k)}$ at the corresponding timestep $t$, the group action $\asn_t^{(k)}$, the
corresponding log-probability, and the critic value
$\valf(\tree_t^{(k)})$.
The profit $\rew_t$ obtained after solving the resulting MPC problem is shared
by all group decisions that contributed to the tree built at time~$t$.

For a group action, the factorized policy in \cref{eq:groupfactorization} gives
\begin{equation}
    \log \policy(\asn_t^{(k)} \mid \tree_t^{(k)})
    =
    \sum_{i\in\mathcal{G}_t^{(k)}}
    \log \policy(\asn_{t}^{(i, k)} \mid \tree_t^{(k)}).
    \label{eq:grouplogprob}
\end{equation}
In practice, we store the mean per-scenario log-probability,
\begin{equation}
    \overline{\log \policy}
    \big(\asn_t^{(k)} \mid \tree_t^{(k)}\big)
    =
    \frac{1}{|\mathcal{G}_t^{(k)}|}
    \sum_{i\in\mathcal{G}_t^{(k)}}
    \log \policy(\asn_{t}^{(i, k)} \mid \tree_t^{(k)}),
    \label{eq:meanlogprob}
\end{equation}
as a scale-normalized surrogate for the group log-probability.

Because the decision at each physical time step optimizes outcomes over a finite
horizon of length $\horizon$, we use $\horizon$-step undiscounted returns to
define the training targets.
This choice both reduces the variance of the critic target by avoiding
long-horizon bootstrapping, and aligns the credit assignment with the horizon
over which actions are optimized for.
For a stored transition $j$  corresponding to the pair $(t_j,k_j)$, we compute
\begin{equation}
  \ret_j
  =
  \sum_{h=0}^{\horizon-1}\rew_{t_j+h},
  \qquad
  \adv_j
  =
  \ret_j - \valf(\tree_{t_j}^{(k_j)}),
  \label{eq:returns}
\end{equation}
truncating the sum at the episode end (\ie when the price data terminates) and standardizing $\adv_j$ within each
batch.
With the probability ratio
\begin{equation}
    \ratio_j
    =
    \frac{
    \policy(\asn_{t_j}^{(k_j)} \mid \tree_{t_j}^{(k_j)})
    }{
    \polold(\asn_{t_j}^{(k_j)} \mid \tree_{t_j}^{(k_j)})
    },
    \label{eq:pporatio}
\end{equation}
as in standard PPO, the actor maximizes the clipped surrogate with an entropy
bonus for exploration, while the critic regresses onto the returns:
\begin{align}
    \ratio_j^{\text{clip}}
    &=
    \clip(\ratio_j, 1-\clipeps, 1+\clipeps), \\
  \mathcal{L}^{\text{actor}}(\theta)
    &=
    -\,\mathbb{E}_j\!\Big[
        \min\big(
          \ratio_j\,\adv_j,\;
          \ratio_j^{\text{clip}}\,\adv_j
        \big)
      \Big] \nonumber \\
    &\quad
    -\;
    \entcoef\,\mathbb{E}_j\!\big[\mathcal{H}(\policy)\big],
  \label{eq:actorloss}\\[2pt]
  \mathcal{L}^{\text{critic}}(\phi)
    &=
    \mathbb{E}_j\!\big[
    \ell_\delta\big(\valf(\tree_{t_j}^{(k_j)}) - \ret_j\big)
    \big],
  \label{eq:criticloss}
\end{align}
where $\ell_\delta(\cdot)$ denotes the Huber loss,
\begin{equation}
    \ell_\delta(x) =
    \begin{cases}
      \frac{1}{2}x^2, & \text{if } |x| \le \delta, \\
      \delta (|x| - \tfrac{1}{2}\delta), & \text{otherwise.}
    \end{cases}
    \label{eq:huber}
\end{equation}

Updates draw minibatches from a small replay buffer of recent rollouts.
This design improves sample efficiency, as collecting new data requires computationally expensive optimization.
At the same time, the buffer size is intentionally kept small, such that samples remain close to the current policy, thereby retaining an approximately on-policy training regime.
We clip gradient norms for numerical stability, and perform early stopping of the inner update loop when the KL divergence between the current policy and the behavior policy $\polold$ exceeds a predefined threshold.
This is a standard practice in PPO to prevent overly large policy updates and ensure that training remains within the trust region implied by the surrogate objective.

\section{Experiment Setup}
\label{sec:setup}
 
\subsection{Synthetic price environment}
\label{sec:prices}
 
We evaluate our method in a synthetic electricity price environment with a
known data-generating process. The purpose is not to reproduce a specific market,
but to create a controlled setting in which uncertainty plays a central role,
allowing us to assess how the quality of the scenario tree affects decisions.

The model combines three key features of electricity prices: persistent
fluctuations, predictable intraday patterns, and rare but significant shocks.
These elements jointly make the scenario generation problem challenging, as both
typical behavior and extreme outcomes must be represented.

We model price dynamics through a latent state variable $z_t$ evolving as
\begin{equation}
  z_{t+1} = z_t + \theta(\mu_t - z_t) + \sigma \varepsilon_t + J_t,
  \quad \varepsilon_t \sim \mathcal{N}(0,1);
  \label{eq:latent}
\end{equation}
where each term corresponds to one of the desired features.

The first component, $\theta(\mu_t - z_t)$, captures mean reversion toward a
time-dependent level $\mu_t$. This level follows a deterministic 24-hour cycle,
representing predictable intraday variation in electricity demand and prices.
The parameter $\theta$ controls how quickly the process returns to this baseline.
The second component, $\sigma \varepsilon_t$, introduces continuous stochastic
fluctuations around the mean. This term generates short-term variability and
ensures that trajectories do not simply track the deterministic cycle.
The final component, $J_t$, models rare but impactful events. These shocks occur
infrequently but can have large magnitudes, with both upward and downward
movements. They are essential to reproduce heavy-tailed behavior, where extreme
price realizations, though rare, have a disproportionate impact.

The observed price is obtained by applying a nonlinear transformation to the
latent state. For moderate values of $z_t$, the mapping is approximately linear,
while extreme values are amplified. As a result, large shocks in the latent state
translate into pronounced price spikes, reinforcing the importance of accurately
capturing tail events in the scenario tree.

From this process we generate two disjoint sets of \emph{price profiles}, one for
training and one for evaluation. Each set contains $200$ trajectories of
$\nsteps = 120$ steps, obtained as independent realizations of
\cref{eq:latent} from a common initial state. Both sets follow
the same distribution; all reported results are computed on the held-out
evaluation set.
 
\subsection{Forecast scenarios}
\label{sec:scenarios}
 
At each control step, the controller is supplied with a fan of $\fan$ price
trajectories over a look-ahead horizon $\horizon \doteq 6$. The fan is generated by
simulating the same process as in \cref{sec:prices}, producing $\fan$ independent trajectories of length $\horizon$ starting from the current state,
each assigned uniform probability $\scprob_i = 1/\fan$. By construction, this forecast is statistically consistent with the underlying dynamics, so that performance differences across methods reflect the quality of the scenario-tree construction rather than forecast bias.

The fan size plays a dual role: it determines both the fidelity of the sampled future distribution and the computational complexity of the aggregation problem. 
In particular, the number of possible assignments grows combinatorially with $\fan$, making both tree construction and policy optimization increasingly expensive as $\fan$ grows. 
For this reason, we train the agent with a relatively small fan, $\fan = 10$, where the multistage program remains computationally tractable and the action space is sufficiently constrained to allow efficient learning.

At evaluation time, we increase the fan size up to $\fan = 300$ to assess how well the learned policy generalizes to denser approximations of the future distribution. 
This setting probes whether the policy has learned a scalable assignment strategy, rather than overfitting to a specific fan size. 
Notably, the policy is trained only once with a limited number of scenarios $\fan = 10$ to ease the computational cost of training. The policy is then applied and evaluated unchanged at larger fan sizes.
 
\subsection{Tree-construction methods}
\label{sec:baselines}
 
All methods feed the same multistage program~\cref{eq:mpc}; they differ
only in the tree handed to it. We compare the proposed agent against four classical
constructions and a perfect-foresight reference.
 
\begin{itemize}
  \item \textbf{RL Agent} (proposed). The learned policy of
    \cref{sec:method}, assigning the $\fan$ scenarios over a fixed topology
    with $\nleaves = 6$ leaves.
  \item \textbf{Oracle}. The dispatch~\eqref{eq:mpc} solved on a single trajectory
    equal to the \emph{realized} future prices. It uses information unavailable at
    decision time and so provides an unattainable upper bound on profit.
  \item \textbf{Random}. The same fixed $6$-leaf topology as the RL Agent, but with
    each scenario assigned to a leaf uniformly at random. This isolates the value of
    the learned assignment from that of the topology itself.
  \item \textbf{Forward}. A scenario tree built by forward selection, the standard
    scenario-reduction heuristic that greedily retains the most representative
    scenarios, at a leaf budget of $6$.
  \item \textbf{Backward}. A scenario tree built by backward reduction, the
    complementary reduction heuristic that successively removes the least
    representative scenarios, at a leaf budget of $6$.
  \item \textbf{Deterministic}. The element-wise mean of the $\fan$ scenarios used
    as a single forecast trajectory, i.e. certainty-equivalent control.
\end{itemize}
 
The RL Agent and Random share an identical, predefined topology, whereas Forward
and Backward construct their own branching at the same leaf budget; Oracle and
Deterministic reduce to a single trajectory. We make this distinction explicit so
that the comparison between learned and random assignment (shared topology) is read
separately from the comparison against the reduction heuristics (different
topology, equal budget).
 
\subsection{Metrics}
\label{sec:metrics}
 
We report three families of metrics, averaged over the evaluation profiles.
 
\begin{itemize}
  \item \textbf{Profit.} The realized cumulative arbitrage profit of a control
    episode, $\sum_{t} \rew_t$, our primary measure of control quality. We report
    its mean and standard deviation across the test set profiles. The Oracle and Deterministic
    methods bracket this metric, serving respectively as the perfect-foresight
    upper bound and the certainty-equivalent reference.
  \item \textbf{Tree size.} The number of nodes $|\Nodes|$ and the effective number
    of leaves $|\Leaves|$ of the constructed tree (after empty nodes are pruned),
    which characterize the complexity of the resulting program and how much
    branching each method actually uses.
  \item \textbf{Computation.} The wall-clock time to build the tree and the
    wall-clock time to solve the program~\eqref{eq:mpc}, per control step, which
    together determine the online cost of each method.
\end{itemize}

\section{Results}
\label{sec:results}
 
We evaluate every method on the same held-out set of $200$ profiles at seven
forecast-fan sizes ranging from $\fan = 10$ to $\fan = 300$. The policy is
trained once, at $\fan = 10$, and applied unchanged at all larger fans, so the
$\fan > 10$ results also test how the learned construction transfers to denser
forecasts than it saw in training. The Oracle ignores the fan and optimizes against
the realized prices, so its score is identical across fan sizes ($3702 \pm 849$) and
serves as a fixed upper bound; the Deterministic baseline is certainty-equivalent
control.
 
\subsection{Profit across fan sizes}
\label{sec:res-profit}
 
\begin{figure*}
  \centering
  \includegraphics[width=\linewidth]{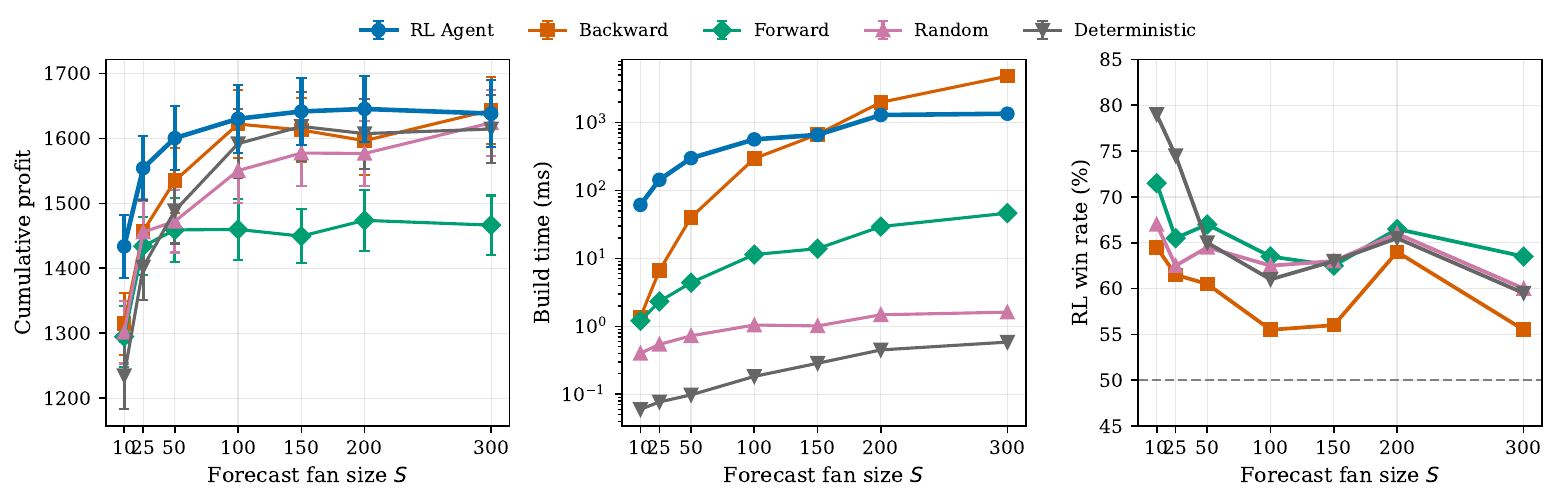}
  \caption{Summary across fan sizes. \emph{Left:} mean profit with error bars
    (non-oracle methods; the Oracle at $3702$ is off-scale). \emph{Centre:}
    per-step tree-build time (log scale). \emph{Right:} the RL Agent's win rate
    against each baseline, with the $50\%$ reference dashed.}
  \label{fig:summary}
\end{figure*}

\begin{figure}
  \centering
  \includegraphics[width=\linewidth]{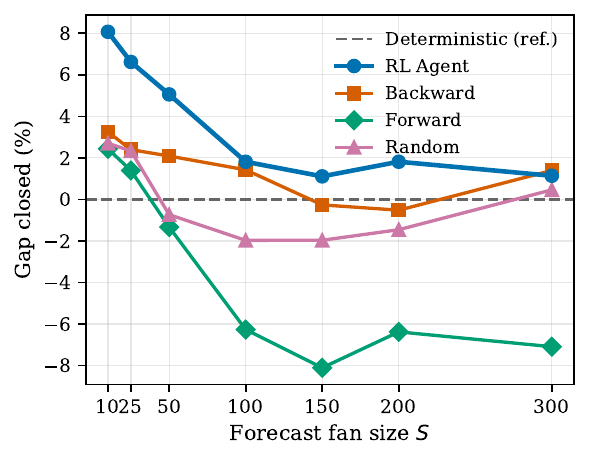}
  \caption{Gap closed versus fan size. The dashed line at $0\%$ is
    certainty-equivalent control (Deterministic) and $100\%$ is the Oracle. The RL
    Agent stays positive at every fan size; the distance-based reductions fall below
    the Deterministic reference at intermediate fan sizes, and Forward remains well below
    it from $\fan=50$ onwards.}
  \label{fig:gap}
\end{figure}

\Cref{tab:profit} reports the mean profits, and the left panel of
\cref{fig:summary} plots it against fan size. To compare methods on a common
scale we report the \emph{gap closed}\,---\,\ie the percentage of the
distance between certainty-equivalent control and the perfect-foresight oracle that
a method recovers\,---\,shown in \cref{fig:gap}.
 
\begin{table*}
  \centering
  \caption{Realized cumulative profit (mean\,$\pm$\,std over the $200$ test profiles).
    Best non-oracle entry per fan size in bold.}
  \label{tab:profit}
  \small
  \setlength{\tabcolsep}{4pt}
  \begin{tabular}{lccccccc}
    \toprule
    & \multicolumn{7}{c}{Forecast fan size $\fan$}\\
    \cmidrule(lr){2-8}
    Method & $10$ & $25$ & $50$ & $100$ & $150$ & $200$ & $300$\\
    \midrule
    Oracle        & $3702\pm849$ & $3702\pm849$ & $3702\pm849$ & $3702\pm849$ & $3702\pm849$ & $3702\pm849$ & $3702\pm849$\\
    \midrule
    RL Agent      & $\mathbf{1434\pm684}$ & $\mathbf{1554\pm696}$ & $\mathbf{1601\pm693}$ & $\mathbf{1630\pm738}$ & $\mathbf{1642\pm730}$ & $\mathbf{1645\pm714}$ & $1638\pm730$\\
    Backward      & $1315\pm675$ & $1457\pm670$ & $1535\pm711$ & $1622\pm747$ & $1613\pm703$ & $1597\pm738$ & $\mathbf{1644\pm731}$\\
    Deterministic & $1235\pm734$ & $1402\pm720$ & $1489\pm713$ & $1592\pm759$ & $1619\pm758$ & $1607\pm763$ & $1615\pm736$\\
    Random        & $1301\pm679$ & $1456\pm722$ & $1472\pm679$ & $1551\pm712$ & $1578\pm715$ & $1577\pm712$ & $1624\pm718$\\
    Forward       & $1295\pm661$ & $1434\pm636$ & $1459\pm704$ & $1460\pm663$ & $1450\pm594$ & $1474\pm668$ & $1467\pm645$\\
    \bottomrule
  \end{tabular}
\end{table*}
 
Most methods (all except forward reduction) converge toward a close value as the fan grows: the profit curves flatten and their error bars overlap heavily at the larger fans (\cref{fig:summary}, left), and the small between-fan fluctuations of the gap-closed curves in this regime (\cref{fig:gap}) are most parsimoniously read as evaluation noise around that common limit. 
The RL Agent reaches it the fastest, with its largest advantage at the smallest fans: it has the highest mean profit at every fan size up to $\fan=200$, by a margin that declines from about $8\%$ of the gap closed at $\fan=10$ to about $2\%$ at $\fan=200$, and at $\fan=300$ backward reduction is marginally higher ($1644$ versus $1638$) within the overlapping error bars. 
The Deterministic baseline improves steadily with the fan and converges for about $\fan=150$ onward (\cref{tab:profit}), from $1235$ at $\fan=10$ to about $1615$ at $\fan\geq150$. 
Indeed, averaging more trajectories resolves the conditional mean price more accurately, and because the dispatch problem is linear, certainty-equivalent control approaches optimality once that mean is well resolved.

For these large fan sizes ($\fan>100$), all but the Forward method converged to each other. Still, even within the convergence region, the RL Agent retains a place advantage: backward reduction's mean at $\fan=300$ ($1644$) does not exceed the RL Agent's at $\fan=200$ ($1645$), so the learned policy reaches the same performance level with roughly a third fewer scenarios and a correspondingly smaller online program. 
More consistently, the RL Agent is the only method whose gap closed stays non-negative across the entire sweep (\cref{fig:gap}): backward reduction and random assignment dip just below the Deterministic reference at intermediate fans ($\fan=150$--$200$) before recovering, and Forward selection sits well below it from $\fan=50$ onward (closing about $-8$ to $-12\%$). Backward reduction and Forward selection behave differently relative to the Deterministic reference. 
Backward's gap closed stays close to it across the whole sweep (between $-0.8\%$ and $+3.0\%$), and the small dips below zero at $\fan=150$ and $\fan=200$ are within the convergence-region noise discussed above. 
Forward selection, by contrast, sits clearly and persistently below the reference from $\fan=50$ onward (between $-2\%$ and $-12\%$), a gap too large and too sustained to attribute to noise. 
A plausible reason is its selection rule: Forward greedily picks scenarios that are far from those already chosen, which favors spread over representativeness. Hence, as the fan grows, the six retained leaves drift toward atypical trajectories rather than tracking the bulk of the forecast. 
The RL Agent, which adapts how many of its six leaves are actually populated (\cref{sec:res-structure}), does not show either pattern.

Random assignment is more surprising. It uses no information about which scenarios
should be retained or grouped\,---\,yet its gap closed stays close to the Deterministic
reference across the sweep (between $-3\%$ and $+3\%$, returning to the reference at
$\fan=300$), and on the lower tail it does better than the mean forecast at every
fan size: $\cvar_{10\%}$ for Random beats Deterministic throughout and is comparable
to Backward's (\cref{tab:cvar}). A plausible reading is that drawing six
scenarios uniformly from the fan is, in expectation, an unbiased sample of the
forecast distribution\,---\,the six retained leaves are then plausible futures rather
than selected extremes or representatives, so the MPC is forced to hedge across
realistic variability instead of committing to the conditional mean, without being
pulled toward atypical trajectories by a selection criterion. The variance inherent
in random sampling thus appears to function as a form of implicit hedging, which is
most useful on the lower tail and consistent with the risk-averse objective that the
controller optimizes.
 
\subsection{Per-profile comparison}
\label{sec:res-winrate}
 
\begin{figure*}
  \centering
  \includegraphics[width=0.75\linewidth]{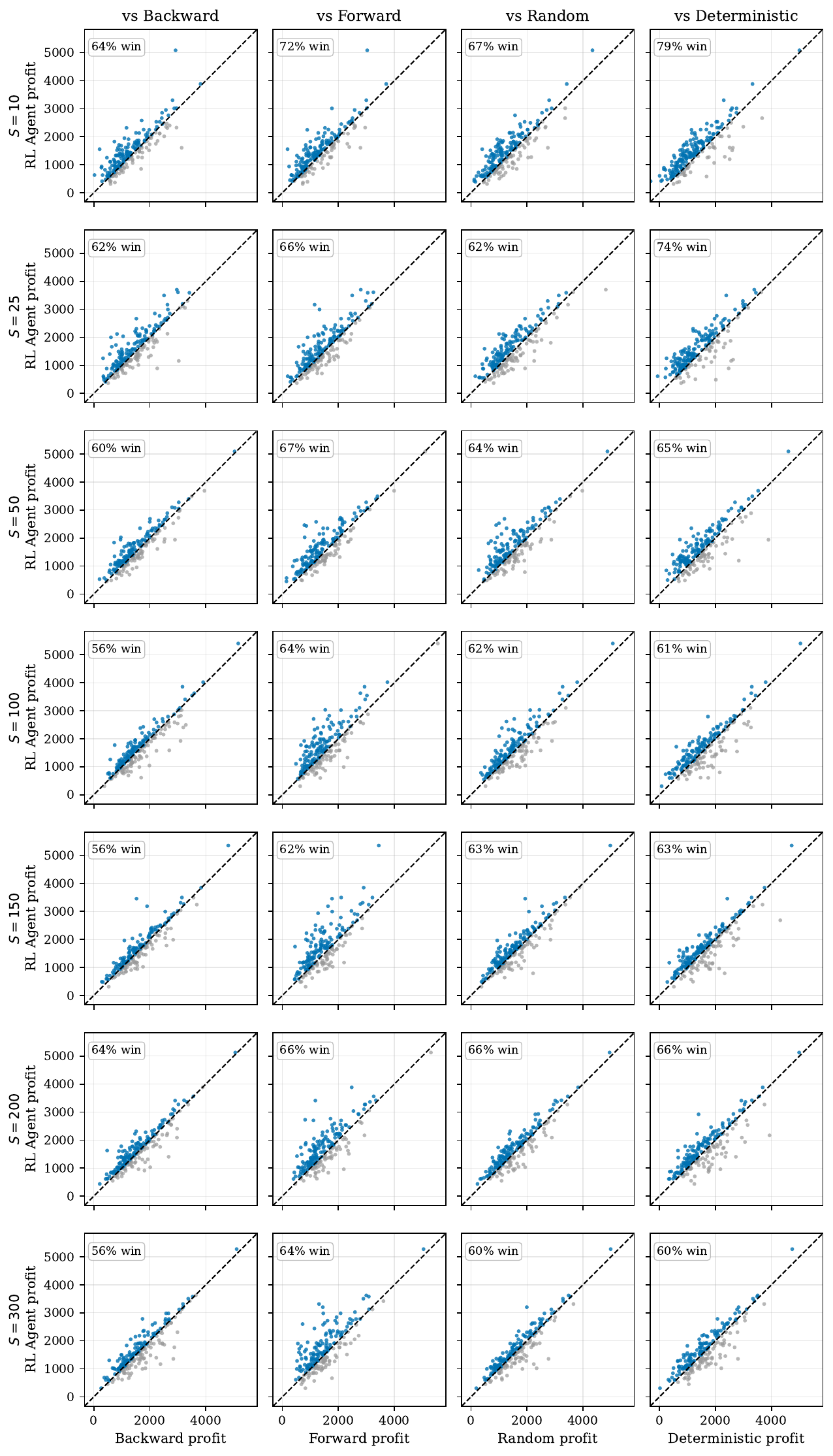}
  \caption{Per-profile profit, RL Agent (vertical) against each baseline
    (horizontal), one row per fan size. Points above the dashed diagonal are
    profiles on which the RL Agent earns more; the boxed figure is the win rate.}
  \label{fig:scatter}
\end{figure*}

The comparison also holds per profile, not only on average.
\Cref{fig:scatter} plots the RL Agent against each baseline profile by
profile: each point is one profile, and points above the diagonal are those on which
the agent earns more. The win rate\,---\,the fraction above the diagonal\,---\,exceeds $50\%$
in every panel, \ie against every baseline at every fan size. Its dependence on the
fan is summarized in the right panel of \cref{fig:summary}: the win rate is
highest against Deterministic at the smallest fan ($79\%$), is lowest against backward
reduction at intermediate fans (about $55$--$56\%$ near $\fan=100$--$150$), and lies
in the $60$--$66\%$ range against all baselines at the larger fans. The win rate
remains above $50\%$ even at $\fan=300$, where the agent and backward reduction have
near-equal mean profit, indicating the agent still wins on a majority of individual
profiles despite the tie in the mean.
 
\subsection{Lower-tail performance}
\label{sec:res-risk}
 
Because the controller optimizes a risk-averse objective, we also compare the
methods on their worst outcomes, using the conditional value-at-risk $\cvar_{5\%}$ and $\cvar_{10\%}$\,---\,\ie the mean profit over the worst $5\%$
and $10\%$ of profiles, respectively. \Cref{tab:cvar} reports
both. On $\cvar_{10\%}$ the RL Agent has the best lower tail at every fan size. On the
more extreme $\cvar_{5\%}$ it is best at every fan size except $\fan=100$ and
$\fan=300$, where Forward selection\,---\,which has a markedly lower mean
(\cref{tab:profit})\,---\,attains a slightly higher value. Read together with
\cref{sec:res-profit}, this gives a sharper statement than either metric alone:
at the larger fans, where the RL Agent's mean profit is at most tied with the best
classical method, its lower tail is clearly better, so the learned policy delivers
the same or higher expected return as the classical reductions while exposing the
controller to substantially less downside. Its tail advantage therefore does not
come from trading expected return for safety; rather, the two improvements coexist.
 
\begin{table}[t]
  \centering
  \caption{Lower-tail profit by method and fan size: $\cvar_{5\%}$ and $\cvar_{10\%}$
    (mean profit over the worst $5\%$ and $10\%$ of profiles). Best non-oracle entry
    per column in bold.}
  \label{tab:cvar}
  \small
  \setlength{\tabcolsep}{4pt}
  \resizebox{\columnwidth}{!}{
  \begin{tabular}{llccccccc}
    \toprule
    & & \multicolumn{7}{c}{Forecast fan size $\fan$}\\
    \cmidrule(lr){3-9}
    & Method & $10$ & $25$ & $50$ & $100$ & $150$ & $200$ & $300$\\
    \midrule
    \multirow{6}{*}{$\cvar_{5\%}$}
      & Oracle        & $2299$ & $2299$ & $2299$ & $2299$ & $2299$ & $2299$ & $2299$\\
      & RL Agent      & $\mathbf{446}$ & $\mathbf{516}$ & $\mathbf{592}$ & $589$ & $\mathbf{568}$ & $\mathbf{633}$ & $559$\\
      & Backward      & $315$ & $423$ & $515$ & $560$ & $497$ & $499$ & $500$\\
      & Deterministic & $121$ & $335$ & $462$ & $372$ & $491$ & $480$ & $449$\\
      & Random        & $276$ & $390$ & $528$ & $504$ & $496$ & $513$ & $544$\\
      & Forward       & $334$ & $405$ & $404$ & $\mathbf{615}$ & $531$ & $532$ & $\mathbf{572}$\\
    \midrule
    \multirow{6}{*}{$\cvar_{10\%}$}
      & Oracle        & $2428$ & $2428$ & $2428$ & $2428$ & $2428$ & $2428$ & $2428$\\
      & RL Agent      & $\mathbf{524}$ & $\mathbf{611}$ & $\mathbf{691}$ & $\mathbf{679}$ & $\mathbf{683}$ & $\mathbf{721}$ & $\mathbf{678}$\\
      & Backward      & $427$ & $533$ & $639$ & $672$ & $630$ & $630$ & $654$\\
      & Deterministic & $278$ & $450$ & $576$ & $515$ & $599$ & $598$ & $585$\\
      & Random        & $382$ & $525$ & $615$ & $594$ & $620$ & $632$ & $653$\\
      & Forward       & $431$ & $530$ & $535$ & $669$ & $626$ & $617$ & $629$\\
    \bottomrule
  \end{tabular}
  }
\end{table}
 
\begin{figure}
  \centering
  \includegraphics[width=\linewidth]{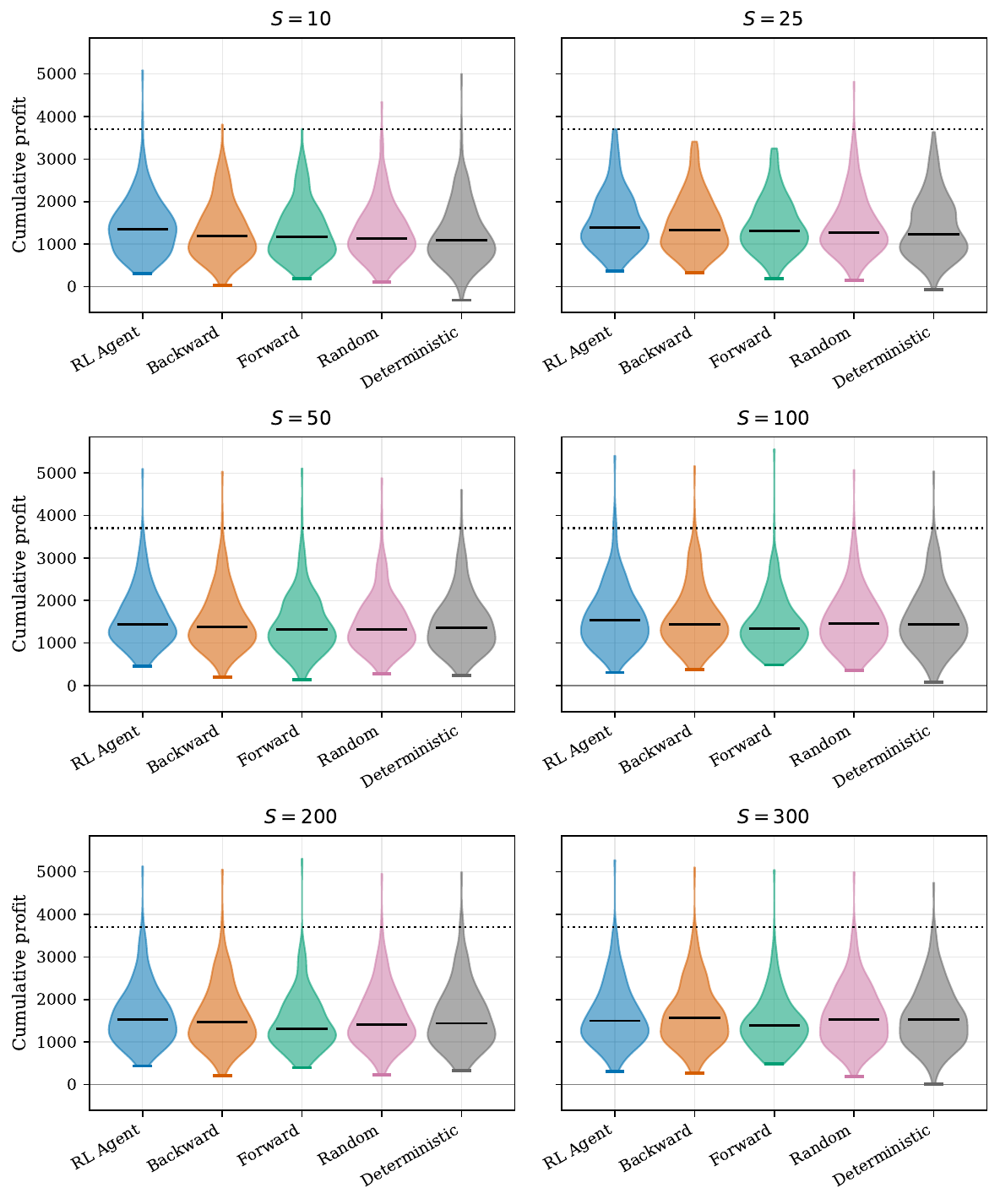}
  \caption{Per-profile profit distributions by method, one panel per fan size.
    Black bars mark medians; coloured ticks mark each method's worst (minimum)
    profile; the dotted line is the Oracle mean. The RL Agent's minimum stays
    positive at all fan sizes, while Deterministic dips to or below zero at the
    smallest and largest fans.}
  \label{fig:dist}
\end{figure}

The difference is clearest at small fans, where the per-profile minima marked in
\cref{fig:dist} show the RL Agent staying profitable on every profile (its
worst-case profit is $308$--$453$ across all fan sizes), while certainty-equivalent
control reaches near-zero or negative worst-case profit at several fan sizes (e.g.\
$-317$ at $\fan=10$, $-66$ at $\fan=25$, and $17$ at $\fan=300$). A likely explanation is that committing to a single mean forecast occasionally mishandles the worst cases, whereas even a little branching hedges against them.
 
\subsection{Tree structure and computation}
\label{sec:res-structure}
 
\Cref{tab:struct} relates the methods' behavior to the trees they build.
Although given six leaves, the RL Agent populates only about two of them on average
(after empty branches are pruned), close to the single trajectory of
certainty-equivalent control, whereas the reduction heuristics always use all six.
This is the structural counterpart of the profit results: the learned policy behaves
like certainty-equivalent control with a small, selective amount of branching, which
is consistent with its tracking the Deterministic baseline from above rather than
dropping below it. The compact trees also keep its LP-solve time roughly half that of
the six-leaf trees.
 
\begin{table}[t]
  \centering
  \caption{Constructed-tree statistics, shown as the range over the seven fan sizes:
    effective leaves used (after pruning, out of six), per-step LP-solve time, and
    per-step tree-build time.}
  \label{tab:struct}
  \small
  \begin{tabular}{lccc}
    \toprule
    Method & Effective leaves & Solve (ms) & Build (ms)\\
    \midrule
    RL Agent      & $1.7$--$2.4$ & $3.4$--$4.5$ & $61$--$1347$\\
    Backward      & $6.0$        & $4.9$--$9.5$ & $1.4$--$4806$\\
    Forward       & $6.0$        & $5.0$--$9.7$ & $1.2$--$46$\\
    Random        & $5.0$--$6.0$ & $5.3$--$10.4$ & $0.4$--$1.6$\\
    Deterministic & $1.0$        & $2.2$--$3.1$ & $0.1$--$0.6$\\
    Oracle        & $1.0$        & $1.4$--$3.3$ & $0.1$--$0.5$\\
    \bottomrule
  \end{tabular}
\end{table}
 
The two costs of online use\,---\,building the tree and solving the program\,---\,trade off across methods. 
The central panel of \cref{fig:summary} shows the build time.
For the RL Agent, it grows with the fan and is its dominant cost (from $61$\,ms at $\fan=10$ to about $1.3$\,s at $\fan=300$). Asymptotically, this growth is quadratic in $\fan$, since building one tree requires $\lceil \fan/ \gsize\rceil$ sequential assignment passes each attending over all $\fan$ scenarios; over the tested range the measured
time grows more slowly, as it is still dominated by the fixed overhead of the sequential passes rather than by the attention itself. 
Backward reduction grows faster\,---\,reflecting its pairwise distance computations\,---\,and overtakes the RL Agent's build time beyond $\fan \approx 150$, reaching about $4.8$\,s at $\fan=300$, roughly three to four times the RL Agent's. This matters for scalability in the convergence region of \cref{sec:res-profit}: for larger fan sizes, the RL Agent and backward reduction obtain similar mean profit, but the gap in construction cost widens sharply, so the cheaper way to achieve that performance level\,---\,particularly if the fan is to be increased further\,---\,is the learned policy rather than its distance-based counterpart. 
Build time is also the binding constraint during training, where a tree is constructed at every step of every rollout, which is why we train at $\fan=10$.
 
\subsection{Discussion}
\label{sec:res-discussion}
 
Across the seven fan sizes, the learned construction is most useful where the forecast is sampled sparsely ($\fan$ is lower), obtaining the highest mean profit, the best lower tail, and a clear majority of per-profile wins. 
As the fan grows the methods converge in mean profit, but within this convergence region the learned policy keeps two practical advantages already noted above. 
It reaches a given performance level with substantially fewer scenarios than backward reduction (its mean at $\fan=200$ matches the best classical mean at $\fan=300$, \cref{sec:res-profit}), and it does so at a build cost that grows more slowly than backward reduction's (\cref{sec:res-structure}). Thus, as the fan size is increased further, the cheaper way to operate near the converged performance level is the learned policy rather than its distance-based counterpart. 
The second observation in the convergence region is that the comparison favors the RL Agent more clearly on the tail than on the mean: it has the best $\cvar_{10\%}$ at every fan size and the best $\cvar_{5\%}$ at all but two (\cref{tab:cvar}), and its worst-case profit stays positive across the entire sweep while certainty-equivalent control posts near-zero or negative worst cases at multiple fan sizes (\cref{fig:dist}). 
The learned policy thus delivers the same or higher expected return as the classical reductions while exposing the controller to substantially less downside, exactly what a risk-averse user would ask of a tree-construction method. A consistent reading of these observations is that the policy adapts how many leaves it uses\,---\,staying near the mean forecast and branching only when it helps\,---\,whereas the fixed six-leaf reductions cannot, so their fit to the fan degrades as it grows.
 
Two limitations bound and contextualize these conclusions. First, the forecast used here is correctly specified by construction, which is the most favorable setting for both certainty-equivalent control and for distance-based reduction, since both rely on the forecast distribution being a faithful description of the underlying process.
The learned policy is trained on closed-loop control profit rather than on the forecast itself, so a biased or otherwise imprecise probabilistic forecast\,---\,the case encountered in practice\,---\,would not directly degrade its objective and may give it additional room to compensate for forecast inaccuracies; we would therefore expect the range of fan sizes over which the learned construction adds value to widen in that setting. 
Second, the dispatch problem studied here is linear, which is what allows certainty-equivalent control to approach optimality as the fan size grows; on non-linear problems, Jensen's inequality precludes that the convergence and the value of a well-constructed tree should remain substantial at all fan sizes. 
Together with using substantially larger forecast fans, evaluating the method on misspecified forecasts and on non-linear control problems is the most informative next step: the results presented here position the approach as a promising, control-oriented alternative to distance-based scenario reduction, but should be regarded as a preliminary investigation whose strongest case will be made on those harder settings.

\section{Conclusions}
\label{sec:conclusions}
 
We have presented a control-oriented, data-driven method for constructing the scenario trees that drive multistage stochastic MPC. 
Departing from the classical view of scenario reduction as an approximation of the forecast distribution under a probability metric, we cast tree construction as the sequential assignment of sampled scenarios to the leaves of a fixed topology and learn it with reinforcement learning, using the realized closed-loop control cost as the sole training signal. 
The tree construction is carried out by an attention-based policy over the scenario set, trained with PPO and supported by an asymmetric critic that exploits privileged future information only during training, so that the deployed controller depends on nothing but the forecast.
 
On a risk-averse battery-arbitrage task evaluated across seven forecast fan sizes from $\fan=10$ to $\fan=300$, the learned construction obtains the highest mean profit at every fan size up to $\fan=200$, with the methods converging at larger fans as certainty-equivalent control approaches optimality on this linear problem. 
Two practical advantages persist throughout the convergence region: our learned RL policy reaches the converged performance level with fewer scenarios and a lower construction cost than backward reduction (which exceeds the agent's
build time beyond $\fan\approx 150$), and its lower tail is consistently better\,---\,the best $\cvar_{10\%}$ at every fan size, a positive worst-case profit throughout\,---\,so the same or higher expected return comes with substantially less downside. 
An analysis of the constructed trees shows that our policy builds compact, near-deterministic trees, populating only about two of its six available leaves on average.
 
Two features of the experimental setting bound these conclusions. 
The dispatch problem is linear, which is what allows certainty-equivalent control to approach optimality as the fan grows; under non-linearities, Jensen's inequality precludes that convergence and the value of a well-constructed tree should remain substantial throughout. 
The forecast is also correctly specified, the most favorable setting for both certainty-equivalent control and distance-based reduction, since both rely on the forecast being a faithful description of the underlying process\,---\,whereas our learned policy is trained on closed-loop profit rather than on the forecast itself, so a biased or imprecise forecast would not directly degrade its objective and may give it additional room to compensate. 
Last, the sizes of the forecasts used to build the tree in the experiment presented are limited to the computational costs of the techniques analyzed.
We therefore position this work as a preliminary investigation whose strongest case will be made on misspecified forecasts, non-linear control problems, and at larger fan sizes, the last reachable with induced-set or linear-attention encoders that reduce the asymptotically quadratic build cost.

\section*{Acknowledgments and funding}
This research was partly funded by the Flemish Government through
the “Onderzoeksprogramma Artificiële Intelligentie (AI) Vlaanderen” programme as well as a travel grant from the Research Foundation - Flanders.

\printcredits

\bibliography{bibl}

@book{morales2014integrating,
  title={Integrating Renewables in Electricity Markets: Operational Problems},
  author={Morales, Juan M and Conejo, Antonio J and Madsen, Henrik and Pinson, Pierre and Zugno, Marco},
  volume={205},
  year={2014},
  publisher={Springer}
}

@book{rawlings2017mpc,
  title={Model Predictive Control: Theory, Computation, and Design},
  author={Rawlings, James B and Mayne, David Q and Diehl, Moritz},
  edition={2},
  year={2017},
  publisher={Nob Hill Publishing}
}

@book{birge2011,
  title={Introduction to Stochastic Programming},
  author={Birge, John R and Louveaux, Fran{\c{c}}ois},
  edition={2},
  year={2011},
  publisher={Springer}
}

@book{shapiro2021,
  title={Lectures on Stochastic Programming: Modeling and Theory},
  author={Shapiro, Alexander and Dentcheva, Darinka and Ruszczy{\'n}ski, Andrzej},
  edition={3},
  year={2021},
  publisher={SIAM}
}

@article{dupacova2003,
  title={Scenario reduction in stochastic programming: An approach using probability metrics},
  author={Dup{\v{c}}ov{\'a}, Jitka and Gr{\"o}we-Kuska, Nicole and R{\"o}misch, Werner},
  journal={Mathematical Programming},
  volume={95},
  number={3},
  pages={493--511},
  year={2003},
  publisher={Springer}
}

@article{heitsch2009,
  title={Scenario tree modeling for multistage stochastic programs},
  author={Heitsch, Holger and R{\"o}misch, Werner},
  journal={Mathematical Programming},
  volume={118},
  number={2},
  pages={371--406},
  year={2009},
  publisher={Springer}
}

@book{pflug2014,
  title={Multistage Stochastic Optimization},
  author={Pflug, Georg Ch and Pichler, Alois},
  year={2014},
  publisher={Springer}
}

@article{rockafellar2000,
  title={Optimization of conditional value-at-risk},
  author={Rockafellar, R Tyrrell and Uryasev, Stanislav},
  journal={Journal of Risk},
  volume={2},
  pages={21--42},
  year={2000},
  publisher={Incisive Media}
}

@inproceedings{donti2017,
  title={Task-based end-to-end model learning in stochastic optimization},
  author={Donti, Priya L and Amos, Brandon and Kolter, J Zico},
  booktitle={Advances in Neural Information Processing Systems},
  volume={30},
  year={2017}
}

@article{elmachtoub2022,
  title={Smart ``predict, then optimize''},
  author={Elmachtoub, Adam N and Grigas, Paul},
  journal={Management Science},
  volume={68},
  number={1},
  pages={9--26},
  year={2022},
  publisher={INFORMS}
}

@article{fairbrother2022,
  title={Problem-driven scenario generation: an analysis of the quality of scenario reduction methods},
  author={Fairbrother, Jamie and Turner, Amanda and Wallace, Stein W},
  journal={Mathematical Programming},
  volume={191},
  number={2},
  pages={141--182},
  year={2022},
  publisher={Springer}
}

@inproceedings{lee2019set,
  title={Set transformer: A framework for attention-based permutation-invariant neural networks},
  author={Lee, Juho and Lee, Yoonho and Kim, Jungtaek and Kosiorek, Adam and Choi, Seungjin and Teh, Yee Whye},
  booktitle={International Conference on Machine Learning},
  pages={3744--3753},
  year={2019},
  organization={PMLR}
}

@article{rujeerapaiboon2022,
  title={Scenario reduction revisited: Fundamental limits and guarantees},
  author={Rujeerapaiboon, Napat and Schindler, Kilian and Kuhn, Daniel and Wiesemann, Wolfram},
  journal={Mathematical Programming},
  volume={191},
  number={1},
  pages={207--242},
  year={2022},
  publisher={Springer}
}

@inproceedings{amos2017optnet,
  title={OptNet: Differentiable optimization as a layer in neural networks},
  author={Amos, Brandon and Kolter, J Zico},
  booktitle={International Conference on Machine Learning},
  pages={136--145},
  year={2017},
  organization={PMLR}
}

@inproceedings{wilder2019,
  title={Melding the data-decisions pipeline: Decision-focused learning for combinatorial optimization},
  author={Wilder, Bryan and Dilkina, Bistra and Tambe, Milind},
  booktitle={Proceedings of the AAAI Conference on Artificial Intelligence},
  volume={33},
  number={01},
  pages={1658--1665},
  year={2019}
}

@inproceedings{gasse2019exact,
  title={Exact combinatorial optimization with graph convolutional neural networks},
  author={Gasse, Maxime and Ch{\'e}telat, Didier and Ferroni, Nicola and Charlin, Laurent and Lodi, Andrea},
  booktitle={Advances in Neural Information Processing Systems},
  volume={32},
  year={2019}
}

@inproceedings{vinyals2015pointer,
  title={Pointer networks},
  author={Vinyals, Oriol and Fortunato, Meire and Jaitly, Navdeep},
  booktitle={Advances in Neural Information Processing Systems},
  volume={28},
  year={2015}
}

@inproceedings{kool2019attention,
  title={Attention, learn to solve routing problems!},
  author={Kool, Wouter and van Hoof, Herke and Welling, Max},
  booktitle={International Conference on Learning Representations},
  year={2019}
}

@article{gros2020data,
  title={Data-driven economic NMPC using reinforcement learning},
  author={Gros, S{\'e}bastien and Zanon, Mario},
  journal={IEEE Transactions on Automatic Control},
  volume={65},
  number={2},
  pages={636--648},
  year={2020},
  publisher={IEEE}
}

@article{cao2020deep,
  title={Deep reinforcement learning-based energy storage arbitrage with accurate lithium-ion battery degradation model},
  author={Cao, Jun and Harrold, Dan and Fan, Zhong and Morstyn, Thomas and Healey, David and Li, Kang},
  journal={IEEE Transactions on Smart Grid},
  volume={11},
  number={5},
  pages={4513--4521},
  year={2020},
  publisher={IEEE}
}

@article{madahi2024distributional,
  title={Distributional reinforcement learning-based energy arbitrage strategies in imbalance settlement mechanism},
  author={Madahi, Seyed Soroush Karimi and Claessens, Bert and Develder, Chris},
  journal={Journal of Energy Storage},
  volume={104},
  pages={114377},
  year={2024},
  publisher={Elsevier}
}

@article{madahi2025model,
  title={Model Predictive Control-Guided Reinforcement Learning for Implicit Balancing},
  author={Madahi, Seyed Soroush Karimi and Bruninx, Kenneth and Claessens, Bert and Develder, Chris},
  journal={arXiv preprint arXiv:2510.04868},
  year={2025}
}

@article{schulman2017proximal,
  title={Proximal policy optimization algorithms},
  author={Schulman, John and Wolski, Filip and Dhariwal, Prafulla and Radford, Alec and Klimov, Oleg},
  journal={arXiv preprint arXiv:1707.06347},
  year={2017}
}

@article{mesbah2016stochastic,
  author  = {Mesbah, Ali},
  title   = {Stochastic Model Predictive Control: An Overview and Perspectives for Future Research},
  journal = {IEEE Control Systems Magazine},
  volume  = {36},
  number  = {6},
  pages   = {30--44},
  year    = {2016},
  doi     = {10.1109/MCS.2016.2602087}
}

@article{zhuang2025problem,
  title={Problem-Driven Scenario Reduction Framework for Power System Stochastic Operation},
  author={Zhuang, Yingrui and Cheng, Lin and Qi, Ning and Almassalkhi, Mads R. and Liu, Feng},
  journal={IEEE Transactions on Power Systems},
  volume={40},
  number={4},
  pages={3232--3246},
  year={2025},
  publisher={IEEE},
  doi={10.1109/TPWRS.2024.3523220}
}

@inproceedings{zhuang2026iterative,
  title={An Iterative Problem-Driven Scenario Reduction Framework for Stochastic Optimization with Conditional Value-at-Risk},
  author={Zhuang, Yingrui and Cheng, Lin and Qi, Ning and Almassalkhi, Mads R. and Liu, Feng},
  booktitle={Proceedings of the 24th Power Systems Computation Conference (PSCC)},
  address={Limassol, Cyprus},
  year={2026}
}

@article{bertsimas2023optimization,
  title={Optimization-based scenario reduction for data-driven two-stage stochastic optimization},
  author={Bertsimas, Dimitris and Mundru, Nishanth},
  journal={Operations Research},
  volume={71},
  number={4},
  pages={1343--1361},
  year={2023},
  publisher={INFORMS},
  doi={10.1287/opre.2022.2265}
}

@article{hewitt2022decision,
  title={Decision-based scenario clustering for decision-making under uncertainty},
  author={Hewitt, Mike and Ortmann, Janosch and Rei, Walter},
  journal={Annals of Operations Research},
  volume={315},
  number={2},
  pages={747--771},
  year={2022},
  publisher={Springer}
}

@article{keutchayan2023problem,
  title={Problem-driven scenario clustering in stochastic optimization},
  author={Keutchayan, Julien and Ortmann, Janosch and Rei, Walter},
  journal={Computational Management Science},
  volume={20},
  number={1},
  pages={13},
  year={2023},
  publisher={Springer}
}

@article{garcia2014iterative,
  title={Iterative scenario based reduction technique for stochastic optimization using conditional value-at-risk},
  author={Garc{\'\i}a-Bertrand, Raquel and M{\'\i}nguez, Roberto},
  journal={Optimization and Engineering},
  volume={15},
  number={2},
  pages={355--380},
  year={2014},
  publisher={Springer}
}

@article{teichgraeber2019clustering,
  title={Clustering methods to find representative periods for the optimization of energy systems: An initial framework and comparison},
  author={Teichgraeber, Holger and Brandt, Adam R},
  journal={Applied energy},
  volume={239},
  pages={1283--1293},
  year={2019},
  publisher={Elsevier}
}

\appendix

\end{document}